\documentclass{egpubl}
\usepackage{eg2024}

\ConferencePaper        %

\usepackage[T1]{fontenc}
\usepackage{dfadobe}  
\usepackage[table]{xcolor}%

\usepackage{amsmath}
\usepackage{cite}  %
\BibtexOrBiblatex
\electronicVersion
\PrintedOrElectronic
\ifpdf \usepackage[pdftex]{graphicx} \pdfcompresslevel=9
\else \usepackage[dvips]{graphicx} \fi

\usepackage{lineno}

\usepackage{xcolor}

\title[TRIPS]%
      {TRIPS: Trilinear Point Splatting for Real-Time Radiance Field Rendering}

\author[L. Franke, D. Rückert, L. Fink \& M. Stamminger]{\parbox{\textwidth}{\centering Linus Franke \orcid{0000-0001-8180-0963}, Darius Rückert \orcid{0000-0001-8593-3974}, Laura Fink \orcid{0009-0007-8950-1790} and Marc Stamminger \orcid{0000-0001-8699-3442}}\\
{\parbox{\textwidth}{\centering Visual Computing Erlangen, Friedrich-Alexander-Universität Erlangen-Nürnberg, Germany\\
		\{firstname.lastname\}@fau.de}}}

\begin{document}
\teaser{
\vspace{-0mm}\includegraphics[width=0.94\linewidth]{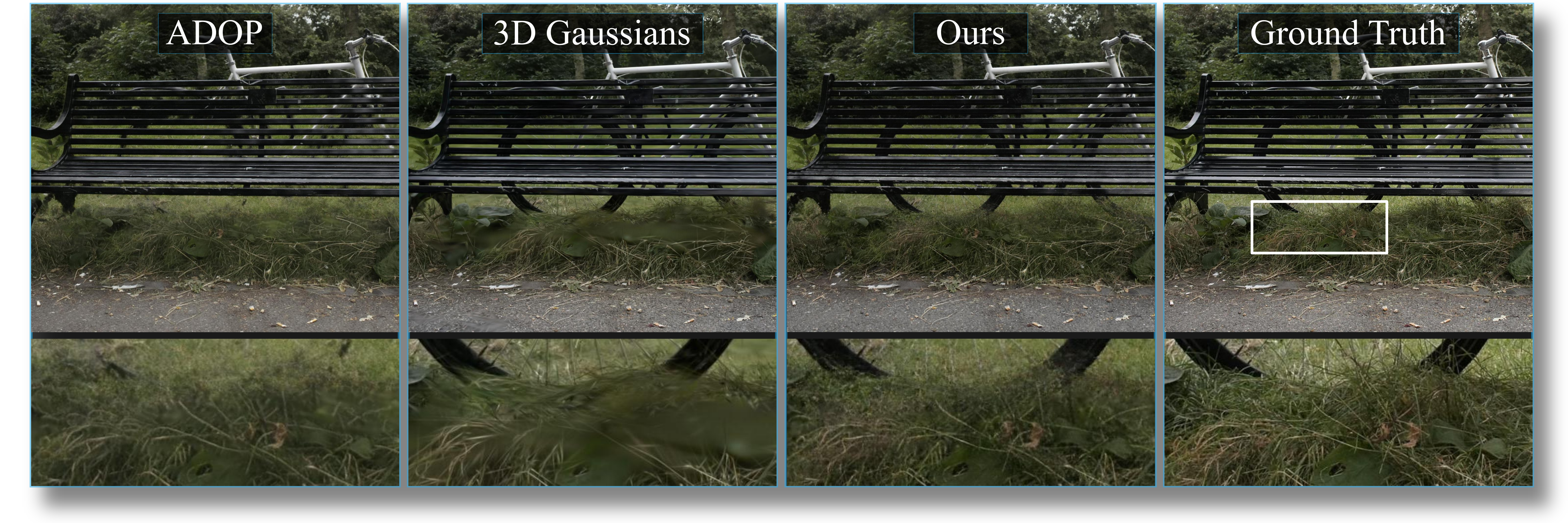}
 \centering\vspace{-0mm}
  \caption{Previous point-based radiance field rendering methods provide great results in many cases, but renderings can be aliased and incomplete (ADOP~\cite{ruckert2022adop} (left), missing parts of the bike's tire), or overblurred (3D Gaussian Splatting~\cite{kerbl20233d} (middle), missing fine grass details). Our approach combines the advantages of both to render crisp, complete, and alias-free images.}
\label{fig:teaser}
}%
\maketitle
\begin{abstract}
Point-based radiance field rendering has demonstrated impressive results for novel view synthesis, offering a compelling blend of rendering quality and computational efficiency. However, also latest approaches in this domain are not without their shortcomings. 3D Gaussian Splatting \cite{kerbl20233d} struggles when tasked with rendering highly detailed scenes, due to blurring and cloudy artifacts. On the other hand, ADOP \cite{ruckert2022adop} can accommodate crisper images, but the neural reconstruction network decreases performance, it grapples with temporal instability and it is unable to effectively address large gaps in the point cloud.

In this paper, we present TRIPS (Trilinear Point Splatting), an approach that combines ideas from both Gaussian Splatting and ADOP.
The fundamental concept behind our novel technique involves rasterizing points into a screen-space image pyramid, with the selection of the pyramid layer determined by the projected point size. 
This approach allows rendering arbitrarily large points using a single trilinear write. 
A lightweight neural network is then used to reconstruct a hole-free image including detail beyond splat resolution.
Importantly, our render pipeline is entirely differentiable, allowing for automatic optimization of both point sizes and positions.

Our evaluation demonstrate that TRIPS surpasses existing state-of-the-art methods in terms of rendering quality while maintaining a real-time frame rate of 60 frames per second on readily available hardware. This performance extends to challenging scenarios, such as scenes featuring intricate geometry, expansive landscapes, and auto-exposed footage.

The project page is located at: \url{https://lfranke.github.io/trips}

\end{abstract}%

\section{Introduction}
Novel view synthesis methods have been a significant driver for computer graphics and vision, 
as they have revolutionized the way we perceive and interact with 3D scenes. 
Many of these methods rely on explicit representations, such as meshes or points. 
Typically, the explicit models are derived from 3D reconstruction processes and can be efficiently rendered through rasterization, which aligns well with contemporary GPU capabilities. 
Nevertheless, these reconstructed models often fall short of perfection and necessitate additional steps to mitigate artifacts.

A common strategy to handle these artifacts is to use scene-specific optimization methods, known as inverse rendering. 
This allows for the adjustment of the scene's texture, geometry, and camera parameters to align the rendering with the photograph. 
Prominent techniques in this domain incorporate per-point descriptors~\cite{ruckert2022adop, aliev2020neural, kopanas2021point}, explicit optimization of point sizes via Gaussians~\cite{kerbl20233d} and learned neural refinement networks~\cite{thies2019deferred, ruckert2022adop, kopanas2022neural}. %
While this generally extends render times, it significantly enhances visual quality.

In the realm of point-based inverse and neural rendering techniques, two successful recent approaches are \textit{3D Gaussian Splatting}\cite{kerbl20233d} and \textit{ADOP}\cite{ruckert2022adop}. 
The former method employs a unique strategy where each point is rendered as a 3D Gaussian distribution, allowing for direct optimization of the points' shape and size.
This process effectively fills gaps in point clouds within the global coordinate space through the utilization of large splats. 
Remarkably, this approach yields high-quality images without necessitating the integration of a neural network for reconstruction. 
However, a drawback is the potential loss of sharpness, as Gaussians tend to introduce blurriness and cloudy artifacts, particularly when there are limited observations available.

In contrast, ADOP rasterizes radiance fields as one-pixel points with depth testing at multiple resolutions. 
Subsequently, it employs a neural network to address gaps and enhance texture details in screen space.
This approach possesses the capability to reconstruct texture details that surpass the resolution of the original point cloud, although the neural network adds an additional computational overhead and shows weaknesses in filling large holes. 

In this paper, we introduce TRIPS, a novel approach that seeks to harness the strengths of both ADOP and 3D Gaussians without loosing real-time rendering capabilities.
Similar to 3D Gaussian Splatting, TRIPS rasterizes splats of varying size, however, like ADOP, it also applies a reconstruction network to generate hole-free and crisp images.
More precisely, we first rasterize the point cloud as $2\times2\times2$ trilinear splats into an image pyramid and blend them using front-to-back alpha blending.
Subsequently, we feed the image pyramid through a compact and efficient neural reconstruction network, which harmonizes the various layers, addresses remaining gaps, and conceals rendering artifacts.
To ensure the preservation of high levels of detail, particularly in challenging input scenarios, we incorporate spherical harmonics and a tone mapping module into our pipeline.

In our evaluations, we demonstrate that our approach can yield crisper images compared to 3D Gaussians, with almost the same perfomance.
Furthermore, it surpasses ADOP in the task of filling sizable gaps and maintaining temporal consistency throughout the rendering process.
In summary, our contributions are:
\begin{itemize}
    \item The introduction of TRIPS, a novel trilinear point splatting technique for radiance field rendering.
    \item A differentiable pipeline for optimization of all input parameters, including point positions and sizes, creating a robust scene representations.
    \item An implementation of the method resulting in high-quality real-time renderings under varying capturing conditions at:\begin{center}
        \url{https://github.com/lfranke/TRIPS}
    \end{center}
 \end{itemize}

\begin{figure*}
	\centering
	\includegraphics[width=.99\linewidth]{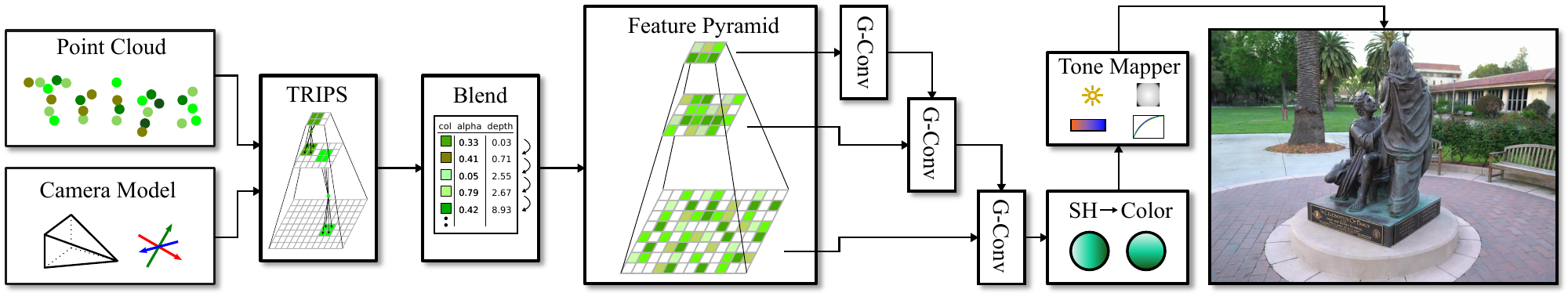}
    \centering
	\caption[]{\label{fig:pipeline} Our pipeline: TRIPS renders and blends a point cloud trilinearly as 2$\times$2$\times$2 splats into multi-layered feature maps with the results being passed though our small neural network, containing only a single gated convolution per layer. Following, an optional spherical harmonics module and tone-mapper is used to produce the final image. This pipeline is completely differentiable, so that point descriptors (colors) and positions, as well as camera parameters are optimized via gradient descent.
    }%
\end{figure*}

\section{Related Work}
In this section, we provide an overview of the field of novel view synthesis and choices for scene representations in this problem domain.

\subsection{Novel View Synthesis and Traditional Approaches}
Traditionally, real-world novel view synthesis relies on image-based rendering techniques.
Commonly, \textit{Structure-from-Motion~(SfM)} techniques~\cite{snavely2006photo,schonberger2016structure} allow camera parameter estimations from a set of photographs which are then used for directly warping source image colors to a target view~\cite{debevec1998efficient,chaurasia2013depth}.
This relies on accurate proxy geometry (usually point clouds or meshed), commonly enhanced via \textit{Multi-View Stereo~(MVS)}~\cite{schoenberger2016mvs,goesele2007multi}. 
In real world datasets however, these techniques can present camera miscalibrations and erroneous geometry~\cite{shum2000review}. 
For image-based rendering, this can lead to warping artifacts, especially near object boundaries, or can cause blurring of details.
Recently, pipelines enhanced by neural rendering~\cite{tewari2022advances} provided powerful tools to lessen these artifacts.

\subsection{Neural Rendering and Scene Representations}
In the last years, multiple variants of deep learning for novel view synthesis were introduced.
Within proxy-based pipelines, several works have replaced the blending operation by a deep neural networks~\cite{riegler2021stable,hedman2018deep,riegler2020free,fink2023livenvs} or learned textures~\cite{thies2019deferred} during the warping stage.
Other approaches use multi-plane images~\cite{mildenhall2019local,srinivasan2019pushing,tucker2020single,zhou2018stereo} or estimate a warping fields~\cite{flynn2016deepstereo,ganin2016deepwarp,zhou2016view} to avoid the need of an scene specific proxy geometry.

This led the way for volumetric scene representations~\cite{penner2017soft} enhanced with deep learning~\cite{sitzmann2020implicit,sitzmann2019deepvoxels} and rendered via ray-marching.
\textit{Neural Radiance Fields (NeRFs)}~\cite{mildenhall2021nerf} furthermore showed that compressing a full 3D scene into a Multilayer Perceptron (MLP) achieve great results in this regard.
This representation however is challenging in its own right, which follow-up works improve upon: long training times~\cite{chen2021mvsnerf,chibane2021stereo, muller2022instant,tancik2021learned,turki2022mega}, many well distributed input views~\cite{chibane2021stereo,yu2021pixelnerf,kopanas2023improving} and rendering times~\cite{muller2022instant,barron2021mipnerf,neff2021donerf}.
Improvements in quality~\cite{barron2023zip,martin2021nerf} allow NeRFs to surpass visual quality of many proxy-based approaches, however render times are still challenging, e.g. \textit{MipNeRF-360}~\cite{barron2022mip} ranging in the order of seconds per image and training needing dozens of hours.

Lately, discretizing parts of the scene space~\cite{yu2021plenoctrees,hedman2021baking} or even replacing parts of it via voxel grids~\cite{Fridovich2022}, octrees~\cite{ruckert2022neat} or tensor factorization~\cite{chen2022tensorf} shrink computational costs as MLPs can be smaller or even removed.
In this area, \textit{InstantNGP}~\cite{muller2022instant} made waves as it uses hash-grids and a highly optimized MLP implementation for faster rendering and trainings speeds while retaining many qualitative advantages of NeRFs.

For the scope of real-time radiance field rendering however, Kerbl and Kopanas et al.~\cite{kerbl20233d} argue that ray-marching as a rendering concept is challenging on current GPU hardware.

\subsection{Real-Time Rendering for Radiance Fields via Points}

In the domain of real-time radiance field rendering, point clouds as an explicit proxy representation remain a great option. 
Point clouds are easily captured via LiDAR-based mapping~\cite{liao2022kitti}, RGB-D cameras with fusion techniques~\cite{dai2017bundlefusion,whelan2016elasticfusion, keller2013realtime} and SfM/MVS techniques~\cite{schoenberger2016mvs}.
They represent a unstructured set of samples in space, with varying distances to neighbors, but true to the originally captured data.
Rendering these can be very fast~\cite{schutz2021rendering,schutz2022software,schutz2019real}, and augmenting points with neural descriptors~\cite{ruckert2022adop,aliev2020neural,rakhimov2022npbg++,franke2023vet,hahlbohm2023plenopticpoints} or optimized attributes~\cite{kopanas2021point,kopanas2022neural} provide high quality renderings using differentiable point renderers~\cite{wiles2020synsin,yifan2019differentiable} or neural ray-based renderers~\cite{xu2022point,ost2022neural,abou2024particlenerf}.
However, discrete rasterization of points can cause aliasing~\cite{schutz2022software} or overdraw~\cite{ruckert2022adop} if many points are rendered to the same pixel. 

Another problem shared by point rendering techniques is how to fill holes in the unstructured data.
Two main approaches have evolved over the years~\cite{kobbelt2004survey}: splatting (in world-space) and screen-space hole filling. 

In world-space hole-filling, points are represented as oriented discs, often termed "splats" or "surfels", with disc radii precomputed based on point cloud density. 
To reduce artifacts between neighboring splats, these discs can be rendered using Gaussian alpha-masks and combined with a normalizing blend function~\cite{alexa2004point,pfister2000surfels,zwicker2001surface}.
Recent techniques optimize splat sizes~\cite{kerbl20233d,zhang2022differentiable} or improve quality with neural networks~\cite{Yang_2020_CVPR}.
For performance, overdraw poses a major issue as splats tend to overlap a lot. 
Thus, special care has to be taken regarding the amount of splats drawn.
3D Gaussian Splatting~\cite{kerbl20233d} can be considered state of the art in this domain. 
They combine anisotropic Gaussians with a very fast tiled renderer and optimize splat sizes via gradient descent.
However, limiting Gaussian numbers is necessary to avoid performance hits, which in turn can lead to over-blurring of small detailed elements.

The second direction involves screen-space hole-filling, where points, often rendered as tiny splats, are post-processed either through traditional methods~\cite{pintus2011real,marroquim2007efficient,grossman1998point} or using convolutional neural networks (CNNs)~\cite{aliev2020neural,meshry2019neural,song2020deep}. While these techniques bridge large point distances, their need for a large receptive field can result in artifacts or performance issues. A multi-resolution pyramid rendering approach mitigates this by assigning different network layers to varied resolutions\cite{aliev2020neural,rakhimov2022npbg++,harrerfranke2023inovis}, albeit reintroducing overdraw issues at lower layers~\cite{ruckert2022adop}. Notably, ADOP~\cite{ruckert2022adop} excels in screen-space hole-filling, enabling the rendering of hundreds of millions of points for sharp object visualization~\cite{schutz2022software}, but encounters challenges with temporal aliasing and substantial hole-filling.

Our approach aims to take the best of both worlds. 
Using TRIPS, we can render large splats by optimizing their size, but avoid high rasterization costs. 
This allows rendering enormous point clouds and detailed textures, while still being real-time capable without aliasing or temporal instability.

\section{Method}

Fig.~\ref{fig:pipeline} provides an overview of our rendering pipeline. 
The input data consists of images with camera parameters and a dense point cloud, which can be obtained through methods like multi-view stereo~\cite{schoenberger2016mvs} or LiDAR sensing. 
To render a specific view, we project the neural color descriptors of each point into an image pyramid using the TRIPS technique (as detailed in Sec.~\ref{sec:trilinearpointrenderer}) and blend them (Sec.~\ref{sec:blending}). 
Subsequently, a compact neural reconstruction network (described in Sec.~\ref{sec:neuralnetwork}) integrates the layered representation, followed by the application of a spherical harmonics module (discussed in Sec.~\ref{sec:shmodule}) and a tone mapper that transforms the resulting features into RGB colors.

Core to our method is the trilinear point renderer, which splats points bilinearly onto the screen space position as well as linearly to two resolution layers, determined by the projected point size.
Our renderer uses similar nomenclature and is inspired by previous point-rasterizing approaches~\cite{schutz2022software,kopanas2021point,ruckert2022adop}.
The neural image $I$ is the output of the render function $\Phi$
 \begin{equation}
 	\text{I} = \Phi(C, R, t, x, E, s_w, \tau, \alpha),
 	\label{eq:render}
 \end{equation}
where $C$ are the camera intrinsics, $(R,t)$ the extrinsic pose of the target view, $x$ the position of the points, $E$ the optional environment map, $s_w$ the world space size of the points, $\tau$ the neural point descriptors and $\alpha$ the transparency for each point.

In contrast to other approaches, we do not use multiple render passes with progressively smaller resolutions, as this causes severe overdraw in the lower resolution layers.
Instead, we compute the two layers which best match the point's projected size and render it only into these layers as $2\times 2$ splat.
By doing so, we mimic varying splat sizes, although effectively rendering only $2\times 2$-splats.
The layers are then later merged in a small neural reconstruction network (Sec.~\ref{sec:neuralnetwork}) to the final image, resembling the decoder part of a U-Net.

\subsection{Differentiable Trilinear Point Splatting}
\label{sec:trilinearpointrenderer}

\begin{figure}
	\centering
	\includegraphics[width=.99\linewidth]{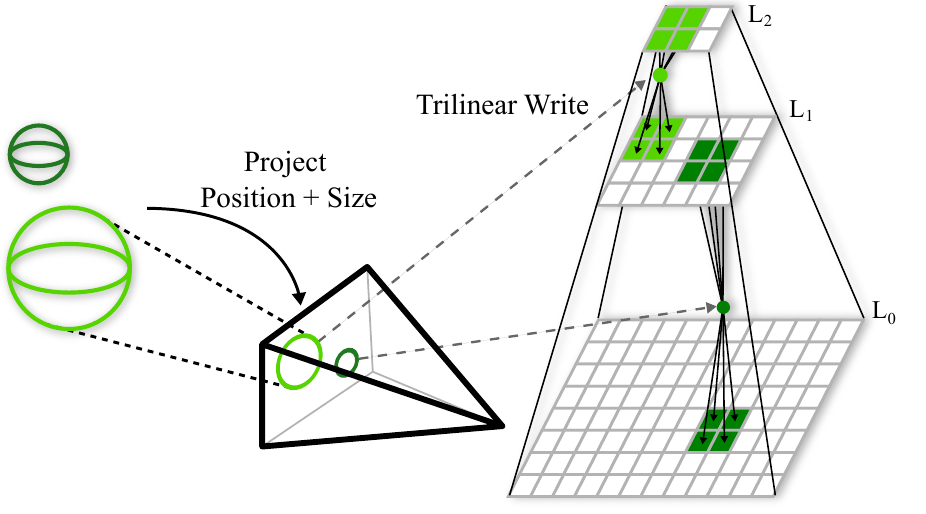}
    \centering
	\caption[]{Trilinear Point Splatting: (left) all points and their respective size are projected into the target image. Based on this screen space size, each point is written to the correct layer of the image pyramid using a trilinear write (right). Large points are written to layers of lower resolution and therefore cover more space in the final image.}
    \label{fig:trilinear}
\end{figure}

Using camera intrinsics $C$ and pose $(R,t)$, we project each point position $(x_w,y_w,z_w)$ to continuous (non-rounded) screen space coordinates $(x,y,z)$ and each world-space point size $s_w$ to screen space size $s$ with the camera's focal length $f$:
\begin{equation}
    s = \frac{f \cdot s_w}{z}.
\end{equation}
Next, we render these points as a $2\times2\times2$ splats bilinearly and handle point size by splatting into two neighboring resolution layers $L$, as shown in Fig.~\ref{fig:trilinear}.
The resolution layers are selected to be the two closest in sizes to the projected size of the point
with $L_{lower} = \lfloor \log(s) \rfloor$ and $L_{upper} =\lceil \log(s) \rceil$.

For each of the then selected eight pixels, we compute the contribution of the point to that pixel and augment its own transparency value value with it.
The final opacity value $\gamma$ that is written to the image pyramid for pixel $(x_i, y_i, s_i)$  with $s_i =  2^L $ 
is
\begin{align}
    \gamma &=   \beta \cdot \iota \cdot \alpha, \\
    \beta &= (1-|x-x_i|) \cdot (1-|y-y_i|) \\
    \iota &= 
	\begin{cases}
		1-\frac{|s - s_i|}{2^{L_{upper}}-2^{L_{lower}}}             & s \geq 1 \\
        \epsilon + (1-\epsilon) s    & s_i = 0 \land s < 1
	\end{cases}
 	\label{eq:alpha_fac}
\end{align}
where $\beta$ is the bilinear weight inside the image layer, $\iota$ is the linear layer weight, and $\alpha$ the opacity value of the point.
The layer weight $\iota$ is a standard linear interpolation if the point size $s$ is inside the image pyramid. 
The second case of Equ.~\eqref{eq:alpha_fac} handles far away points that have a pixel size smaller than one.
In order not to miss these, we always add them to the finest level $0$.
To avoid that their weight disappears, we ensure that their contribution is at least $\epsilon=0.25$.

\subsection{Multi Resolution Alpha Blending}
\label{sec:blending}

\begin{figure}
	\centering
	\includegraphics[width=.80\linewidth]{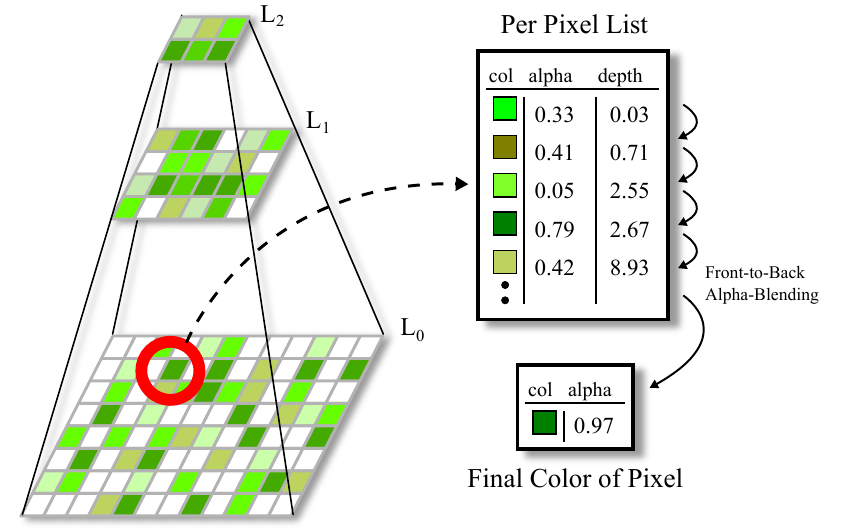}
    \centering
	\caption[]{In each pixel of the image pyramid, a depth-sorted list of colors and alpha values is stored. The final color of each pixel is computed using front-to-back alpha blending on the sorted list.}
    \label{fig:blending}
\end{figure}

Since each point is written to multiple pixels and multiple points can fall into the same pixel, we collect all fragments in per pixel lists $\Lambda_{l_i,x_i,y_i}$. 
These lists are sorted by depth and clamped to a maximum size of 16 elements.
Eventually, the color $C_\Lambda$ is computed using front-to-back alpha blending (Fig.~\ref{fig:blending}):
\begin{equation}
    C_{\Lambda} = \sum_{m=1}^{|\Lambda|} T_m \cdot \alpha_m \cdot c_m 
    \label{eq:blend}
\end{equation} 
\begin{equation}
    T_m =\prod_{i=1}^{m-1} (1-\alpha_i),
\end{equation}

\begin{figure}
	\centering
	\includegraphics[width=.99\linewidth]{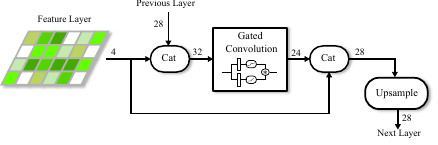}
    \centering
	\caption[]{Our design of one gated convolution block that processes the features of the image pyramid with the number of channels passed through indicated at each step.}
    \label{fig:conv}
\end{figure}

\subsection{Neural Network}
\label{sec:neuralnetwork}

The result produced by our renderer consists of a feature image pyramid comprising $n$ layers. These individual layers are finally consolidated into a single full-resolution image by a compact neural network, as depicted in Fig.~\ref{fig:pipeline}.
Our network architecture incorporates a single gated convolution~\cite{yu2019free} in each layer with a self-bypass connection and a feature size of 32. Additionally, we include a bilinear upsampling operation for all layers except the final one, merging the output with the subsequent level. This configuration is shown in Fig.~\ref{fig:conv} and resembles an efficient decoder network, due to its restrained number of features, pixels, and convolutional operations.

Unlike well-established hole-filling neural networks~\cite{aliev2020neural,ruckert2022adop,rakhimov2022npbg++}, our approach demands a significantly smaller and more efficient network. This reduced network size stems from the fact that our renderer is adept at filling gaps autonomously and generates smooth output through trilinearly splatting points. Consequently, the network's primary task is to learn minimal hole-filling and outlier removal, allowing it to concentrate its efforts on high-quality texture reconstruction.

\subsection{Spherical Harmonics Module and Tone Mapping}
\label{sec:shmodule}
To model view dependent effects and camera-specific capturing parameters (like exposure time), we optionally interpret the network output as spherical harmonics (SH) coefficients, convert them to RGB colors, and finally pass the result to a physically-based tone mapper.
This allows the system to make use of explicit view directions.
The SH-module makes use of spherical harmonics with degree 2, which corresponds to 27 input coefficients (9 coefficients per color channel).
These coefficients are the output of the last convolution of our network.
The tone mapper follows the work of R\"uckert et al.~\cite{ruckert2022adop}, which models exposure time, white balance, sensor response, and vignetting.

\subsection{Optimization Strategy}

Before novel views can be synthesized, the rendering pipeline is optimized to reproduce the input photographs. 
This optimization includes point position, size, and features, as well as the camera model and poses, neural network weights, and tone mapper parameters.
We train for 600 epochs, which, depending on scene size, requires 2-4 hours to converge.

As training criterion, we use the VGG-loss~\cite{johnsonvgg16} which has been shown to provide high-quality results~\cite{ruckert2022adop}.
The VGG network, however, tends to be slow to evaluate, thus increasing training times significantly compared to MSE loss.
Therefore, we use a combination of MSE and SSIM~\cite{kerbl20233d} in the first 50 epochs when the advantages of VGG are still negligible.
This speeds up training time by about $5\%$ percent.

Similar to Kerbl and Kopanas~\cite{kerbl20233d}, we use a "warm-up" period of 20 epochs, during which we train with half image resolutions.
Afterwards we randomly zoom in and out each epoch, so that all convolutions (whose weights are not shared) are trained to contribute to the final result.

\subsection{Implementation Details}
\label{sec:impl_details}
Our implementation uses \textit{torch} as auto-differentiable backend, however the trilinear renderer is implemented in custom CUDA kernels, as they commonly provide better performance~\cite{kerbl20233d,ruckert2022adop}.
Fast spherical harmonics encodings are provided by \textit{tiny-cuda-nn}~\cite{tiny-cuda-nn}.

The renderer is implemented in three stages: collecting, splatting and accumulation, albeit diverging from other state-of-the-art multi-layer blending strategies, this turned out to work best in our scenario~\cite{franke2018multi,Lassner_2021_CVPR, vasilakis2020survey}.
We first project each point $(x_w,y_w,z_w)$ to the desired view and collect each point's $(x,y,z)$ as well as point size $s$ in a buffer, and also count how many elements are mapped to each pixel.
This counting is then used for an offset scan to index into one continuous arrays for all layers.
The following splatting pass duplicates each point and stores a pair of $(z,i)$ (with $i$ an index to the stored information) in each pixels' list.

Following, a combined sorting and accumulation pass is done.
Regarding performance, this part is critical, as such we opt to only use the front most 16 elements from each (sorted) list, a common practice when blending points~\cite{Lassner_2021_CVPR}. 
We could not identify any loss of quality caused by this approximation, as the blending contribution of later points is very low.
This limitation allows us to use GPU-friendly sorting, as we repeat warp-local (32 threads) and shuffle-based bitonic sorts, always replacing the latter 16 elements with new unsorted ones, until the lists are empty.
For the backwards pass, the sorted per-pixel lists are stored, allowing fast backpropagation.
The front-to-back alpha blending (see Sec.~\ref{sec:blending}) is done in the same pass as the sorting pass, because all relevant elements are already in registers. 

In contrast to Kerbl and Kopanas et al.~\cite{kerbl20233d}, we use this per-pixel sorting, which proved to be faster for us then global sorting. This is mostly due to the higher amount and smaller sizes of points in our approach.

For scenes with a large deviation in point density, we found that occlusion may not be correctly evaluated by the neural network in edge cases.
Therefore, we include points from coarser layers during blending (in the usual way), of which the additional cost is very small~($<0.5$ms).

Point sizes are initialized with the average distance to the four nearest neighbor, which is then efficiently optimized during training (see Fig.~\ref{fig:horse_pointsize}).

\definecolor{best}{rgb}{0.53,0.74,0.27}
\definecolor{second}{rgb}{0.93,0.88,0.36}
\definecolor{third}{rgb}{0.93,0.75,0.2}
\begin{table*}[ht]
\centering
\caption[]{Results on the Tanks\&Temples and MipNeRF-360 datasets, as well as \textsc{Boat} and \textsc{Office}. See also Fig.~\ref{fig:big_table_fig} for visual comparisons.}%
\small\begin{tabular}{@{}l|ccc|ccc|ccc|ccc@{}}%
  & \multicolumn{3}{c|}{\textit{Tanks\&Temples}} & \multicolumn{3}{c|}{\textit{MipNeRF-360}}& \multicolumn{3}{c|}{\textsc{Boat}}   & \multicolumn{3}{c}{\textsc{Office}} \\
\textit{Method}        &LPIPS$\downarrow$& PSNR$\uparrow$ & SSIM$\uparrow$ &  LPIPS$\downarrow$  & PSNR$\uparrow$   & SSIM$\uparrow$ &  LPIPS$\downarrow$  & PSNR$\uparrow$ & SSIM$\uparrow$   &  LPIPS$\downarrow$  & PSNR$\uparrow$ & SSIM$\uparrow$\\ \hline
InstantNGP   & 0.475 & 21.74 & 0.692                                                                                & 0.374 & \cellcolor{third}25.94 & 0.697                                                                & 0.598 & 15.34 & 0.455                                                                                 & 0.544 & 13.45 & 0.801 \\
Mip-NeRF 360 & 0.340 & \cellcolor{third}24.61 & 0.789                                                               & 0.286 &  \cellcolor{best}\textbf{28.23} &  \cellcolor{best}\textbf{0.796}                             & 0.680 & 12.20 & 0.357                                                                                 & 0.526  & 15.22  & 0.832 \\
Gaussian Spl.& \cellcolor{third}0.300 & \cellcolor{second}\textit{24.63} &  \cellcolor{best}\textbf{0.818}          & \cellcolor{second}{0.278} & \cellcolor{second}\textit{26.94} & \cellcolor{second}\textit{0.792}      & \cellcolor{third}0.544 & \cellcolor{third}15.30 & \cellcolor{third} 0.470                             & \cellcolor{third}0.371 & \cellcolor{third}18.77 & \cellcolor{third}0.878 \\
ADOP         & \cellcolor{second}\textit{0.229} & 23.78 & \cellcolor{third}0.802                                    & \cellcolor{third}\textit{0.285} & 23.26 & 0.707                                                       &  \cellcolor{best}\textbf{0.301} &  \cellcolor{best}\textbf{20.49} &  \cellcolor{best}\textbf{0.650}   & \cellcolor{second}\textit{0.279} &  \cellcolor{best}\textbf{21.47} &  \cellcolor{best}\textbf{0.899} \\
Ours         & \cellcolor{best}\textbf{0.213} &  \cellcolor{best}\textbf{24.64} & \cellcolor{second}\textit{0.808}  &  \cellcolor{best}\textbf{0.233} & \cellcolor{third}25.94 & \cellcolor{third}0.772                     & \cellcolor{best}\textbf{0.301} & \cellcolor{second}\textit{20.38} & \cellcolor{second}\textit{0.633}  & \cellcolor{best}\textbf{0.271} & \cellcolor{second}\textit{21.36} & \cellcolor{second}\textit{0.887} \\
\end{tabular}
\label{tab:eval_qual}%
\end{table*}

\section{Evaluation}
Next, we compare our approach with prior arts as well as showcase the effectiveness of our design decisions in ablation studies.

\subsection{Setup and Datasets}

We have evaluated our approach on several scenes from the Tanks\&Temples~\cite{Knapitsch2017} and the MipNeRF-360~\cite{barron2022mip} datasets.
Additionally, we use the \textsc{Boat} and \textsc{Office} scene from R\"uckert et al.~\cite{ruckert2022adop} to evaluate robustness towards difficult input conditions.
The former contains outdoor auto-exposed images while the later is a office floor with multiple distinct room and a large LiDAR point cloud, but sparsely placed cameras.

From Tanks\&Temples, we use the \textit{intermediate} set containing eight scenes: \textsc{Train}, \textsc{Playground}, \textsc{M60}, \textsc{Lighthouse}, \textsc{Family}, \textsc{Francis}, \textsc{Horse} and \textsc{Panther}.
These scenes are outdoor scenes captured under varying lighting conditions but with good spatial coverage and can be seen as a good baseline for robustness.
The MipNeRF-360 dataset~\cite{barron2022mip} consists of 5 outdoor and 4 indoor scenes.
This dataset was captured with controlled setups and has capture positions well suited for volumetric rendering with a hemispherical setup~\cite{kopanas2023improving}. 
We use half resolution for images of this dataset, resulting in resolutions of around $2500\times1600$ px for outdoor and $1550\times1030$ px for indoor scenes.
For results with the resolutions used in related works (outdoor: quarter resolution; indoor half resolution), take a look at the Appendix, Tabs.~\ref{tab:360_scene_lpips}-\ref{tab:360_scene_ssim}.

Point clouds of all scenes were acquired via COLMAP's MVS~\cite{schoenberger2016mvs}, except \textsc{Office} which was captured by LiDAR.

For the quantitative evaluation we use the LPIPS\textsubscript{VGG}~\cite{zhang2018perceptual}, PSNR, and SSIM metrics. 
We note however, that neither of these metrics always reflect visual impression.
Some approaches are trained with MSE-loss or SSIM and therefore naturally perform better in PSNR and SSIM. 
Our approach, on the other hand, is trained with VGG-loss and thus usually shows better scores on LPIPS.
For a fair comparison, we recommend to look at all metrics and closely inspect the provided image and video comparison.

In all experiments, we leave every 8th view out for testing. This is the same train/test split as used in current related work~\cite{barron2022mip,kerbl20233d}.

\subsection{Quality Comparison}

\begin{figure*}
	\centering
	\includegraphics[width=.99\linewidth]{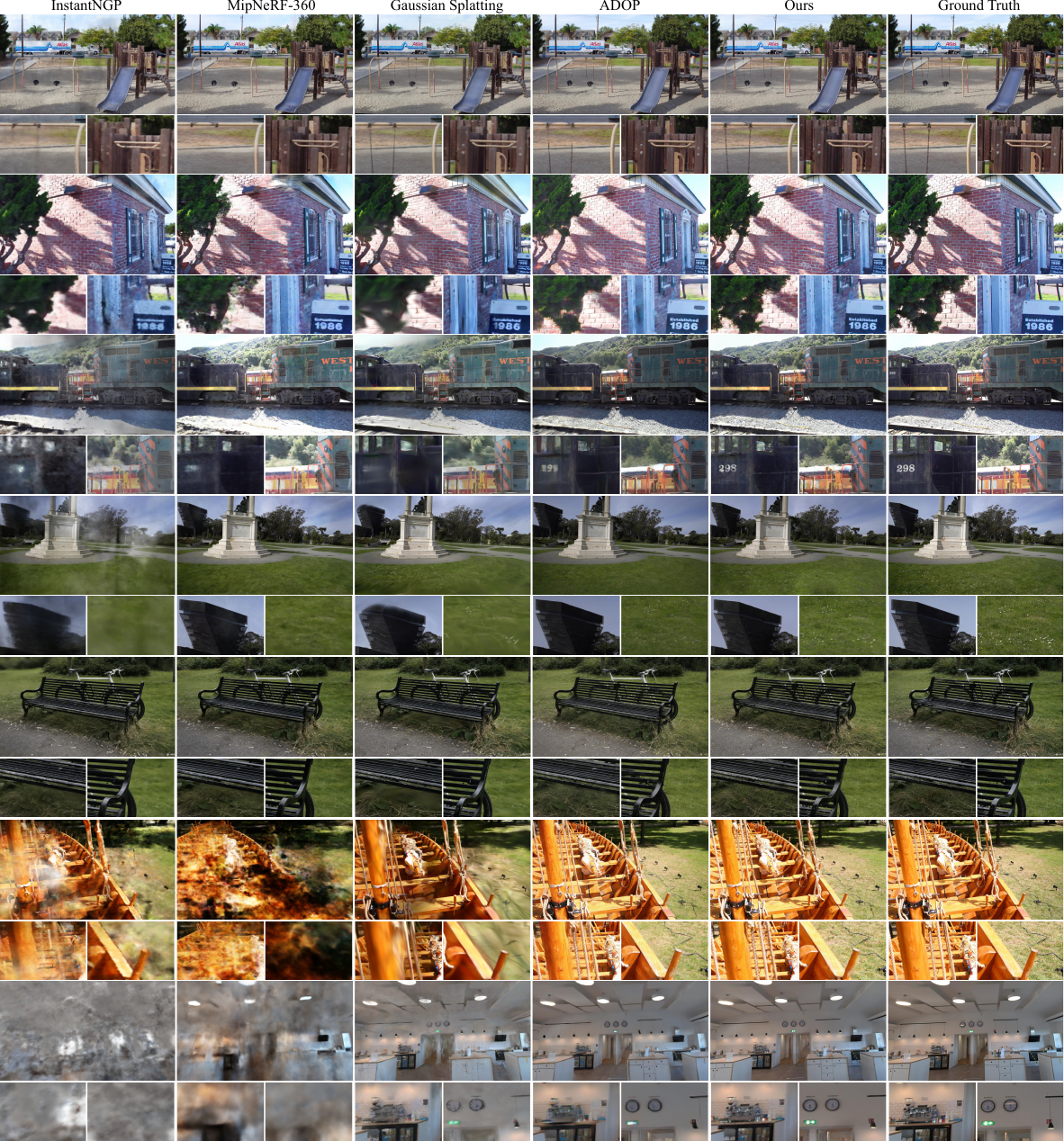}%
	\caption[]{\label{fig:big_table_fig}Visual comparisons. }%
\end{figure*}

In Tab.~\ref{tab:eval_qual} and Fig.~\ref{fig:big_table_fig}, we compare our approach to InstantNGP~\cite{muller2022instant}, MipNeRF-360~\cite{barron2022mip}, 3D Gaussian Splatting~\cite{kerbl20233d} and ADOP~\cite{ruckert2022adop}. 
The latter two are the closest-related point-based radiance field rendering approaches.

On the Tanks\&Temples dataset, our approach achieves in average the best LPIPS score with an improvement of 20\% over the second best.
In PSNR and SSIM the score is on par with state-of-the-art.
On the MipNeRF-360 dataset, we obtain again the best LPIPS score, however the volumetric methods and Gaussian Splatting show an improved PSNR and SSIM.
The difference can be inspected in Fig.~\ref{fig:big_table_fig}. 
For example, in row 3, the TRIPS rendering provides better sharpness with more details, but the MipNeRF-360 and Gaussian output is overall cleaner with less noise.
On the difficult \textsc{boat} and \textsc{office} scenes, we can show that our rendering pipeline, is robust to extreme input conditions.

Individual scores per scene can be seen in the Appendix in Tabs.~\ref{tab:tt_scene_lpips}-\ref{tab:own369_scene_ssim}.
Video results are showcased at \url{https://youtu.be/Nw4A1tIcErQ}.

\subsection{Ablation Studies}
In this section, first we show the effect of our design choices.

\subsubsection{Point-Size Optimization}
\begin{figure}
	\centering
	\includegraphics[width=.99\linewidth]{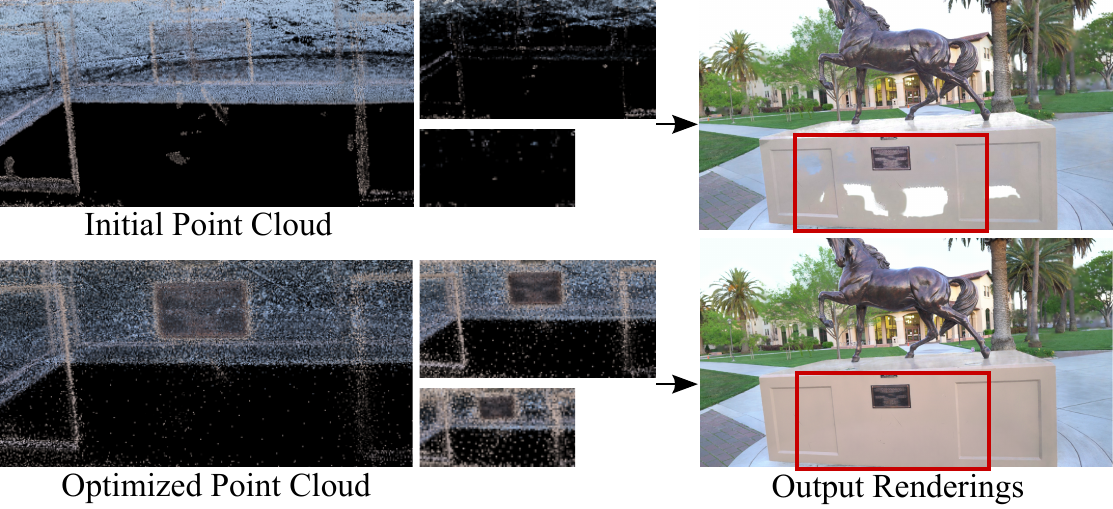}
    \centering
	\caption[]{ 
The initial COLMAP reconstruction lacks points on the pedestal of the statue (top left). Our approach distributes the few present points and increases their sizes (bottom left) thus rendering them also in lower layers (middle). Thus our pipeline can avoid distracting holes (right).
 \label{fig:horse_pointsize}
    }%
\end{figure}
With our trilinear splatting technique, point sizes can be optimized to fill large holes in the scene.
We show this capability in Fig.~\ref{fig:horse_pointsize}, where the initial point cloud exhibits a large hole in the pedestal of the horse producing artifacts in rendering (top row).
To combat this, our pipeline efficiently moves and enlarges the points to fill the hole (bottom row), thus providing great render quality.

\begin{table*}
\caption{\label{tab:view_dep} View dependency on different scenes. On scenes with strong view dependency (\textsc{Garden}), adding view dependant configurations, either via our SH network module (\textit{SH-net}) or optimized per point (\textit{SH-point}) increases quality, however the per-point point setup severely impacts performance. Our module gives a balanced trade off, which also avoids over-fitting on less view-dependent scenes (\textsc{Playground}).}%
\footnotesize%
\centering%
\begin{tabular}[b]{l|cccc|cccc|cccc}
& \multicolumn{4}{c|}{\textsc{Playground} (12.5M Points)} & \multicolumn{4}{c|}{\textsc{Horse} (1.8M Points)}  & \multicolumn{4}{c}{\textsc{Garden} (8.2M Points)}\\
\textit{View-dep } & LPIPS  $\downarrow$ & PSNR $\uparrow$  & SSIM $\uparrow$ 
 & Time $\downarrow$& LPIPS  $\downarrow$ & PSNR $\uparrow$  & SSIM $\uparrow$ 
 & Time $\downarrow$& LPIPS  $\downarrow$ & PSNR $\uparrow$  & SSIM $\uparrow$ 
 & Time $\downarrow$\\ \hline 
none    & 0.225 & 24.85 & 0.720 & 11.1ms & 0.202 & 22.73 & 0.822 & 7.7ms & 0.219 & 24.82 & 0.756 & 17.9ms\\
SH-net   & 0.225 & 24.88 & 0.724 & 13.3ms & 0.203 & 22.81 & 0.825 & 8.9ms & 0.222 & 24.46 & 0.752 &  18.5ms   \\
SH-point & 0.236 & 24.32 & 0.702& 27.4ms & 0.200 & 22.89 & 0.829 & 10.2ms & 0.213 & 25.15 & 0.756 & 27.3ms
\end{tabular}%
\end{table*}

\subsubsection{Point Position Optimization}
\begin{figure}
	\centering
	\includegraphics[width=.95\linewidth]{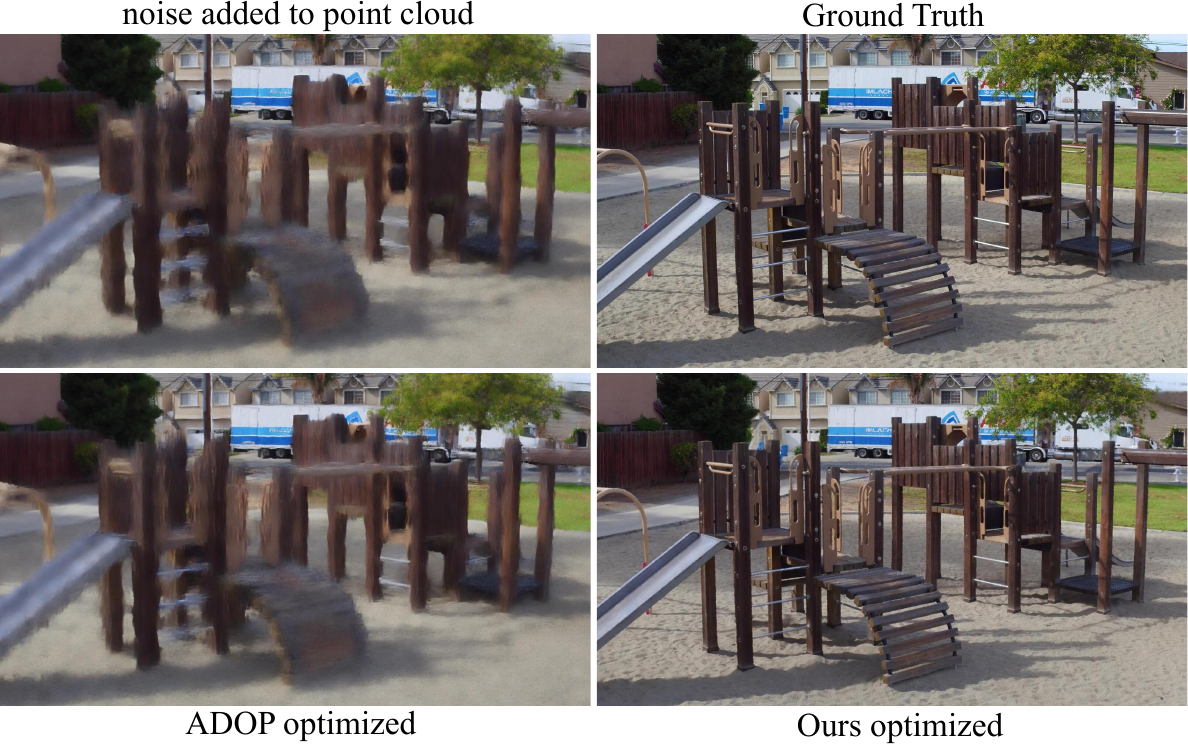}
    \centering
	\caption[]{We added noise to the converged point clouds of ADOP and ours, then restarted optimization for positions only.
    Ours is able to converge back to the correct result, ADOP fails at that.
 \label{fig:point_noise}
    }%
\end{figure}
To test the efficiency of our trilinear point position optimization compared to the (cheaper) approximate gradients from ADOP, we added random noise (of $0.01$) to the positions of all points after training, then optimize only point positions for 100 epochs.
The result can be seen in Fig.~\ref{fig:point_noise}. 
Our pipeline is able to reconstruct the correct rendering, while ADOP's result barely improves.

\begin{table}[]
\caption{\label{tab:layers_num} Number of resolution layers used (\textsc{horse} scene). }\centering%
\footnotesize%
\begin{tabular}[b]{l|cccc}
\textit{\#Layers } & LPIPS  $\downarrow$ & PSNR $\uparrow$  & SSIM $\uparrow$ 
 & Time $\downarrow$\\ \hline 
3        & 0.216              & 21.24    & 0.818 & 7.10ms      \\
4        & 0.201              & 22.40    & 0.826 & 7.35ms      \\
5        & 0.197              & 22.76    & 0.826 & 7.42ms      \\
6        & 0.196              & 23.06    & 0.829 & 7.50ms      \\
7        & 0.195              & 23.10    & 0.826 & 7.54ms      \\
8        & 0.192              & 23.34    & 0.828 & 7.61ms      \\
\end{tabular}
\end{table}
\subsubsection{Number of Render Layers}
Due to our trilinear point rendering algorithm, increasing the number of pyramid layers has almost no negative impact on render time.
As seen in Tab.~\ref{tab:layers_num}, having 8 layers improves quality, especially with PSNR.
For reference, other approaches make use of 4~\cite{ruckert2022adop} or 5~\cite{aliev2020neural} layers and describe significant performance impacts when increasing the number of layers~\cite{ruckert2022adop}.

\subsubsection{View Dependency}
After the neural network, optionally we use a spherical harmonics module to model view depended artifacts of the scene.
This improves the rendering quality for some scenes (\textsc{Garden}), while for others it makes little to no difference (see Tab.~\ref{tab:view_dep}).
Applying the spherical harmonics before the network achieves roughly the same quality, but also reduces efficiency due to additional memory overhead.
On scenes without reflective materials, skipping the spherical harmonics module is thus possible.

\subsubsection{Feature Vector Dimensions}
Our pipeline uses by default four feature descriptors per point.
More features only marginally increase the quality, while requiring significantly more memory and slightly increasing rendering time, as shown in Tab.~\ref{tab:feature_num}.
\begin{table}[]
\caption{\label{tab:feature_num} Features per point on the \textsc{playground} scene. }\centering%
\footnotesize%
\begin{tabular}[b]{l|cccc}%
\textit{\# Features } & LPIPS  $\downarrow$ & PSNR $\uparrow$  & SSIM $\uparrow$ 
 & Time $\downarrow$\\ \hline 
4 & 0.225 & 24.85 & 0.720 & 11.1ms      \\
6 & 0.231 & 24.61 & 0.701 & 11.7ms      \\
8 & 0.223 & 25.04 & 0.727 & 12.2ms      \\
\end{tabular}
\end{table}

\subsubsection{Networks}

In our pipeline, we use a small decoder network made out of gated convolutions, presented in Sec.~\ref{sec:neuralnetwork}.
ADOP~\cite{ruckert2022adop} on the other hand uses a four layer U-net with double convolutions for encoder and decoder (thus around 6 times more parameter).
As seen in Tab.~\ref{tab:networks}, in our pipeline our networks provide similar quality to ADOP's full network, while being much faster in inference.
With spherical harmonics, inference times slightly increase, but the system is now able to model view dependency.
Adding the SH-module to the second finest layer (ours+SH$_{\text{L2}}$) instead of the finest (ours+SH) of the network improves efficiency but weakens results.

\begin{table}[]
\caption{\label{tab:networks}Network configuration compared (\textsc{Playground} scene).}\centering%
\footnotesize%
\begin{tabular}[b]{l|cccc}
\textit{Network } & LPIPS  $\downarrow$ & PSNR $\uparrow$  & SSIM $\uparrow$ 
 & Time $\downarrow$\\ \hline 
ADOP-net    & 0.236        & 24.74    & 0.713 & 10.7ms       \\ 
ours        & 0.225        & 24.85    & 0.720 & 4.5ms      \\
ours+SH     & 0.225            &  24.88       &0.724     & 5.6ms        \\
ours+SH$_{\text{L2}}$  & 0.248            &  24.34       &0.684      & 2.2ms        \\  
\end{tabular}
\end{table}

\subsubsection{Time Scaling on Number of Points}

As seen in Tab.~\ref{tab:point_rend_eff}, TRIPS is very efficient in rendering large amounts of points. 
\begin{table}[h]%
\caption{\label{tab:point_rend_eff} Efficiency of our approach regarding point cloud sizes. }\centering%
\footnotesize%
\begin{tabular}[b]{l|ccccc}%
\textit{Scene }     &\textsc{Horse}& \textsc{Garden} & \textsc{Playgr.} & \textsc{Boat} & \textsc{Office} \\\hline %
\#Points      & 1.8M & 7.8M  & 12.5M & 53.0M& 72.5M      \\
Time    & 2.5ms  & 5.9ms & 6.2ms& 13.1ms& 15.0ms  
\end{tabular}%
\end{table}%
Even for our largest scene with more than 70M points, the pipeline remains real-time capable with only 15ms required for rasterization.

\subsection{Rendering Efficiency}

In Tab.~\ref{tab:eval_time}, we evaluate training and rendering time for all examined methods.
Our method trains for around 2-4h per scene on an Nvidia A100 and renders a novel view in around 11ms on an RTX4090.
A finer breakdown of the steps involved can be found in Tab.~\ref{tab:rendering_breakdown}. 

\begin{table}[]
\caption[]{\label{tab:eval_time}%
Training and render times on the \textsc{Garden} (images resolution: 2594$\times$1681) and \textsc{Playground} scene (1920$\times$1080).  %
}%
\footnotesize\centering%
\begin{tabular}{l|r|r|r} %
\textit{Method} &   \multicolumn{1}{c|}{Train }&  \multicolumn{1}{c|}{Render\textsc{(Garden)}} &  \multicolumn{1}{c}{Render\textsc{(Playgr.)}} \\ \hline
InstantNGP      & 0.25h & 131ms  & 172ms \\
Mip-NeRF360    & 36h & 38000ms & 18000ms \\
ADOP            & 8h & 30.3ms & 14.5ms  \\
Gaussian Spl.   & 0.75h & 11.5ms & 8.6ms  \\
Ours            & 4h & 16.4ms &  11.1ms \\
\end{tabular}%
\end{table}

\begin{table}[]
\caption{\label{tab:rendering_breakdown}Breakdown of the frame time for the \textsc{playground} scene. Our method's "Rasterize" consists of: counting and memory allocation with 1.9ms, splatting with 2.6ms and combined sorting and blending with 1.7ms.}\centering%
\centering
\footnotesize\begin{tabular}{l|r|rrr|r}
\textit{Method}&\#Points & Rasterize & Network & Tonemap & In Total \\\hline
ADOP       & 12M   & 3.1ms    & 11.0ms   & 0.4ms     & 14.5ms    \\
Gauss. Spl.& 2M & 8.6ms    &          &           &  8.6ms   \\
Gauss. Spl.& 8M & 11.4ms & & & 11.4ms \\
Ours      &12M    & 6.2ms    & 4.5ms    & 0.4ms     & 11.1ms    \\
\end{tabular}
\end{table}

\subsection{Outlier Robustness}
\begin{figure}[]
	\centering
	\includegraphics[width=.99\linewidth]{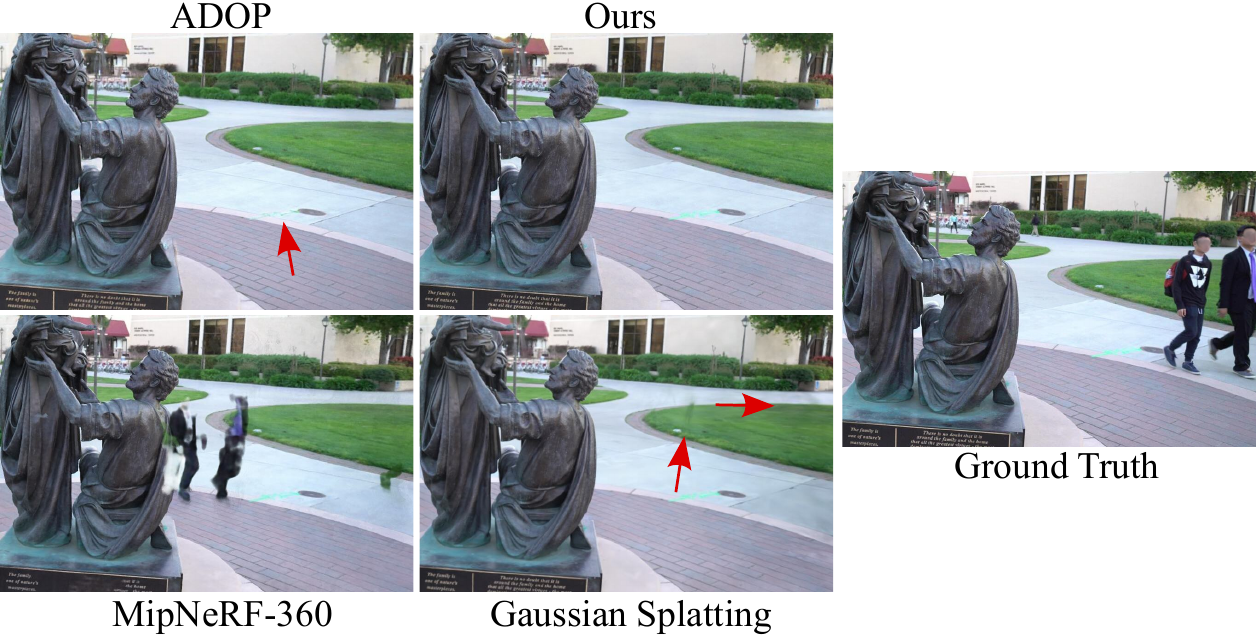}
    \centering
	\caption[]{\label{fig:outlier_robustness}
        Comparison of outlier robustness on the \textsc{family} scene. Only our methods is able to remove floating artifacts while still retaining full color precision on the sidewalk.
    }%
\end{figure}
As seen in Fig.~\ref{fig:outlier_robustness}, our approach is robust to outlier measurements, for example, people walking through the scene.
Especially volumetric approach like MipNeRF-360, suffer from severe artifacts in this case, due to strong view-dependant over-fitting capability.

\subsection{Comparison to Prior Work with Number of Points}

We have seen in previous experiments that Gaussian Splatting~\cite{kerbl20233d} has blurrier results compared to TRIPS, which can be confirmed by their weak LPIPS scores.
However, they start with fewer point primitives (the SfM reconstruction) and thus are limited in the amount of detail to display.
To this end, we conducted an experiment, where the Gaussian Splatting pipeline is provided with the dense point cloud (providing the same input as for our pipeline).
Gaussian splatting has a pruning mechanism to remove unwanted Gaussian, thus after their full training, from the initial 12.5M points only around 8M survived. 

\begin{figure}
	\centering
	\includegraphics[width=.99\linewidth]{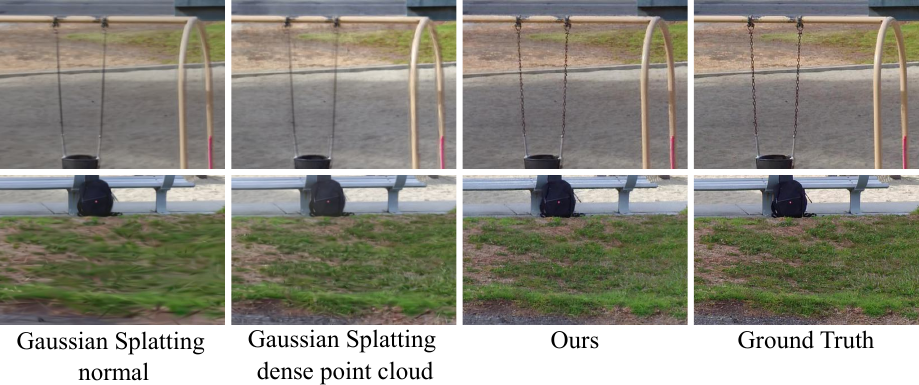}
    \centering
	\caption[]{\label{fig:gaussian_large_small}
        Visual results of Gaussian splatting with COLMAP's dense point cloud as input compared its normal setup as well as ours, which provides the sharpest results (\textsc{Playground} scene).
        }%
\end{figure}

The results of this experiment are presented in Tab.~\ref{tab:performance_large_oc}.
It can be seen that LPIPS improves with more Gaussians (however PSNR declines) as fine details can be reconstructed better.
The qualitative comparison paints the same picture (see Fig.~\ref{fig:gaussian_large_small}), where the quality of the grass improves drastically, however finer details such as the chains still can only be reconstructed by us.
Overall the technique cannot reach the quality and scores of TRIPS, as we can keep more points to render efficiently as well as use neural descriptors to encode more detailed information.

Furthermore, our approach performs more efficiently in scenarios with large point clouds.
In the dense setup, TRIPS outperforms Gaussian Splatting, as the resolution-dependant computation cost of our neural network (4.5ms at $1920\times1080$) catches up with our more efficient point rasterizer (see Tab.~\ref{tab:rendering_breakdown}).

\begin{table}[]
\caption{\label{tab:performance_large_oc} Performance of the methods on the  \textsc{Playground} scene. Gaussian (dense) starts with COLMAP's dense reconstruction of 12M points and prunes them to 8M, Gaussian (sparse) is the original sparse setup and has about 2M points. Also see Fig.~\ref{fig:gaussian_large_small}. }\centering%
\footnotesize%
\begin{tabular}[b]{l|cccc}
\textit{Method } & LPIPS$\downarrow$ & PSNR$\uparrow$  & SSIM$\uparrow$ 
 & Time $\downarrow$\\ \hline 
Ours        & 0.229              & 25.12    & 0.746 & 11.1ms      \\
ADOP        & 0.233              & 24.86    & 0.753 & 14.5ms      \\
Gaussian \textit{(dense)}    & 0.283        & 24.06    & 0.773 & 11.4ms      \\
Gaussian \textit{(sparse)}    & 0.322        & 24.61    & 0.776 & 8.6ms      \\
\end{tabular}
\end{table}

\section{Limitations}
In the preceding section, we have demonstrated TRIPS' effectiveness on commonly encountered real-world datasets. 
Nonetheless, we have also identified potential limitations. 
One such limitation arises from the prerequisite to have an initial dense reconstruction (in contrast to Gaussian Splatting), which may not be practical in certain scenarios. 

\begin{figure}[]
	\centering
	\includegraphics[width=.99\linewidth]{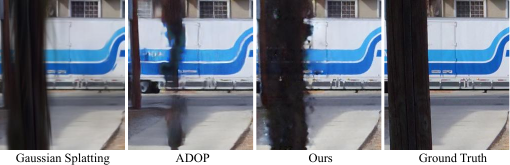}
    \centering
	\caption[]{\label{fig:limitations_fig}
        Limitation: Holefilling close to the camera exhibits fuzzy edges and shine-through.}%
\end{figure}

Additionally, our lack of an anisotropic splat formulation can create problems:
When our method is tasked with strong holefilling of elongated, slender object (such as poles), noisy artifacts surrounding their silhouettes can be observed. 
An example of this is depicted in Fig.~\ref{fig:limitations_fig}.
In such instances, the slightly blurred edge characteristic of Gaussian Splatting is often preferred.

Furthermore, even though the temporal consistency compared to previous point rendering approaches~\cite{aliev2020neural,ruckert2022adop} has been drastically improved, slight flickering can still occur in areas with too many or too little points.

Our trilinear point splatting splits up points into distinct layers and as such looses depth information. 
Theoretically, during recombination this could create holes in solid geometry. 
In practice, we could not find instances of this happening except in extreme zoom-ins far outside the training data.
We believe that the per-point descriptors, the point inclusion in coarse layers, and the network-based recombination are capable to combat this issue, as reflected in the rendering quality.

\section{Conclusion}
In this paper, we presented TRIPS, a robust real-time point-based radiance field rendering pipeline.
TRIPS employs an efficient strategy of rasterizing points into a screen-space image pyramid, allowing the efficient rendering of large points and is completely differentiable, thus allowing automatic optimization of point sizes and positions. 
This technique enables the rendering of highly detailed scenes and the filling of large gaps, all while maintaining a real-time frame rate on commonly available hardware.

We highlight that TRIPS achieves high rendering quality, even in challenging scenarios like scenes with intricate geometry, large-scale environments, and auto-exposed footage. 
Moreover, due to the smooth point rendering approach, a comparably simple neural reconstruction network is sufficient, resulting in real-time rendering performance.

An open source implementation is available under:
{\par\centering\url{https://github.com/lfranke/TRIPS}\par}

\section*{Acknowledgements}

We thank Matthias Innmann, Stefan Romberg, Michael Gerstmayr and Tim Habigt for the fruitful discussions as well as NavVis GmbH for providing the Office dataset.

Linus Franke was supported by the Bayerische Forschungsstiftung (Bavarian Research Foundation) AZ-1422-20.
The authors gratefully acknowledge the scientific support and HPC resources provided by the Erlangen National High Performance Computing Center (NHR@FAU) of the Friedrich-Alexander-Universit\"at Erlangen-N\"urnberg (FAU) under the NHR project \textit{b162dc}. NHR funding is provided by federal and Bavarian state authorities. NHR@FAU hardware is partially funded by the German Research Foundation (DFG) – 440719683.

\bibliographystyle{eg-alpha-doi} 
\bibliography{bib}       

\newcommand{\etalchar}[1]{$^{#1}$}
\begin{thebibliography}{\uppercase{WSMG{\etalchar{*}}16}}

\bibitem[ACDS24]{abou2024particlenerf}
\textsc{Abou-Chakra J., Dayoub F., S{\"u}nderhauf N.}:
\newblock Particlenerf: A particle-based encoding for online neural radiance
  fields.
\newblock In \emph{Proceedings of the IEEE/CVF Winter Conference on
  Applications of Computer Vision} (2024), pp.~5975--5984.

\bibitem[AGP{\etalchar{*}}04]{alexa2004point}
\textsc{Alexa M., Gross M., Pauly M., Pfister H., Stamminger M., Zwicker M.}:
\newblock Point-based computer graphics.
\newblock In \emph{ACM SIGGRAPH 2004 Course Notes}. 2004, pp.~7--es.

\bibitem[ASK{\etalchar{*}}20]{aliev2020neural}
\textsc{Aliev K.-A., Sevastopolsky A., Kolos M., Ulyanov D., Lempitsky V.}:
\newblock Neural point-based graphics.
\newblock In \emph{European Conference on Computer Vision} (2020), Springer,
  pp.~696--712.

\bibitem[BMT{\etalchar{*}}21]{barron2021mipnerf}
\textsc{Barron J.~T., Mildenhall B., Tancik M., Hedman P., Martin-Brualla R.,
  Srinivasan P.~P.}:
\newblock Mip-nerf: A multiscale representation for anti-aliasing neural
  radiance fields.
\newblock In \emph{Proceedings of the IEEE/CVF International Conference on
  Computer Vision (ICCV)} (October 2021), pp.~5855--5864.

\bibitem[BMV{\etalchar{*}}22]{barron2022mip}
\textsc{Barron J.~T., Mildenhall B., Verbin D., Srinivasan P.~P., Hedman P.}:
\newblock Mip-nerf 360: Unbounded anti-aliased neural radiance fields.
\newblock In \emph{Proceedings of the IEEE/CVF Conference on Computer Vision
  and Pattern Recognition} (2022), pp.~5470--5479.

\bibitem[BMV{\etalchar{*}}23]{barron2023zip}
\textsc{Barron J.~T., Mildenhall B., Verbin D., Srinivasan P.~P., Hedman P.}:
\newblock Zip-nerf: Anti-aliased grid-based neural radiance fields.
\newblock \emph{arXiv preprint arXiv:2304.06706} (2023).

\bibitem[CBLPM21]{chibane2021stereo}
\textsc{Chibane J., Bansal A., Lazova V., Pons-Moll G.}:
\newblock Stereo radiance fields (srf): Learning view synthesis for sparse
  views of novel scenes.
\newblock In \emph{Proceedings of the IEEE/CVF Conference on Computer Vision
  and Pattern Recognition} (2021), pp.~7911--7920.

\bibitem[CDSHD13]{chaurasia2013depth}
\textsc{Chaurasia G., Duchene S., Sorkine-Hornung O., Drettakis G.}:
\newblock Depth synthesis and local warps for plausible image-based navigation.
\newblock \emph{ACM Transactions on Graphics (TOG) 32}, 3 (2013), 1--12.

\bibitem[CXG{\etalchar{*}}22]{chen2022tensorf}
\textsc{Chen A., Xu Z., Geiger A., Yu J., Su H.}:
\newblock Tensorf: Tensorial radiance fields.
\newblock In \emph{Computer Vision -- ECCV 2022} (Cham, 2022), Avidan S.,
  Brostow G., Ciss{\'e} M., Farinella G.~M., Hassner T., (Eds.), Springer
  Nature Switzerland, pp.~333--350.

\bibitem[CXZ{\etalchar{*}}21]{chen2021mvsnerf}
\textsc{Chen A., Xu Z., Zhao F., Zhang X., Xiang F., Yu J., Su H.}:
\newblock Mvsnerf: Fast generalizable radiance field reconstruction from
  multi-view stereo.
\newblock In \emph{Proceedings of the IEEE/CVF International Conference on
  Computer Vision} (2021), pp.~14124--14133.

\bibitem[DNZ{\etalchar{*}}17]{dai2017bundlefusion}
\textsc{Dai A., Nie{\ss}ner M., Zollh{\"o}fer M., Izadi S., Theobalt C.}:
\newblock Bundlefusion: Real-time globally consistent 3d reconstruction using
  on-the-fly surface reintegration.
\newblock \emph{ACM Transactions on Graphics (ToG) 36}, 4 (2017), 1.

\bibitem[DYB98]{debevec1998efficient}
\textsc{Debevec P., Yu Y., Boshokov G.}:
\newblock Efficient view-dependent ibr with projective texture-mapping.
\newblock In \emph{EG Rendering Workshop} (1998), vol.~4.

\bibitem[FHSS18]{franke2018multi}
\textsc{Franke L., Hofmann N., Stamminger M., Selgrad K.}:
\newblock Multi-layer depth of field rendering with tiled splatting.
\newblock \emph{Proceedings of the ACM on Computer Graphics and Interactive
  Techniques 1}, 1 (2018), 1--17.

\bibitem[FKYT{\etalchar{*}}22]{Fridovich2022}
\textsc{Fridovich-Keil S., Yu A., Tancik M., Chen Q., Recht B., Kanazawa A.}:
\newblock Plenoxels: Radiance fields without neural networks.
\newblock In \emph{2022 IEEE/CVF Conference on Computer Vision and Pattern
  Recognition (CVPR)} (2022), pp.~5491--5500.

\bibitem[FNPS16]{flynn2016deepstereo}
\textsc{Flynn J., Neulander I., Philbin J., Snavely N.}:
\newblock Deepstereo: Learning to predict new views from the world's imagery.
\newblock In \emph{Proceedings of the IEEE conference on computer vision and
  pattern recognition} (2016), pp.~5515--5524.

\bibitem[FRF{\etalchar{*}}23a]{fink2023livenvs}
\textsc{Fink L., R{\"u}ckert D., Franke L., Keinert J., Stamminger M.}:
\newblock Livenvs: Neural view synthesis on live rgb-d streams.
\newblock In \emph{SIGGRAPH Asia Conference Papers} (New York, NY, USA, Dec.
  2023), Association for Computing Machinery.

\bibitem[FRF{\etalchar{*}}23b]{franke2023vet}
\textsc{Franke L., R{\"u}ckert D., Fink L., Innmann M., Stamminger M.}:
\newblock Vet: Visual error tomography for point cloud completion and
  high-quality neural rendering.
\newblock In \emph{SIGGRAPH Asia Conference Papers} (New York, NY, USA, Dec.
  2023), Association for Computing Machinery.

\bibitem[GD98]{grossman1998point}
\textsc{Grossman J.~P., Dally W.~J.}:
\newblock Point sample rendering.
\newblock In \emph{Eurographics Workshop on Rendering Techniques} (1998),
  Springer, pp.~181--192.

\bibitem[GKSL16]{ganin2016deepwarp}
\textsc{Ganin Y., Kononenko D., Sungatullina D., Lempitsky V.}:
\newblock Deepwarp: Photorealistic image resynthesis for gaze manipulation.
\newblock In \emph{European conference on computer vision} (2016), Springer,
  pp.~311--326.

\bibitem[GSC{\etalchar{*}}07]{goesele2007multi}
\textsc{Goesele M., Snavely N., Curless B., Hoppe H., Seitz S.~M.}:
\newblock Multi-view stereo for community photo collections.
\newblock In \emph{2007 IEEE 11th International Conference on Computer Vision}
  (2007), IEEE, pp.~1--8.

\bibitem[HFF{\etalchar{*}}23]{harrerfranke2023inovis}
\textsc{Harrer M., Franke L., Fink L., Stamminger M., Weyrich T.}:
\newblock Inovis: Instant novel-view synthesis.
\newblock In \emph{SIGGRAPH Asia Conference Papers} (New York, NY, USA, Dec.
  2023), Association for Computing Machinery.

\bibitem[HKT{\etalchar{*}}23]{hahlbohm2023plenopticpoints}
\textsc{Hahlbohm F., Kappel M., Tauscher J.-P., Eisemann M., Magnor M.}:
\newblock Plenopticpoints: Rasterizing neural feature points for high-quality
  novel view synthesis.
\newblock In \emph{Proc. Vision, Modeling and Visualization ({VMV})} (2023),
  Eurographics.

\bibitem[HPP{\etalchar{*}}18]{hedman2018deep}
\textsc{Hedman P., Philip J., Price T., Frahm J.-M., Drettakis G., Brostow G.}:
\newblock Deep blending for free-viewpoint image-based rendering.
\newblock \emph{ACM Transactions on Graphics (TOG) 37}, 6 (2018), 1--15.

\bibitem[HSM{\etalchar{*}}21]{hedman2021baking}
\textsc{Hedman P., Srinivasan P.~P., Mildenhall B., Barron J.~T., Debevec P.}:
\newblock Baking neural radiance fields for real-time view synthesis.
\newblock In \emph{Proceedings of the IEEE/CVF International Conference on
  Computer Vision} (2021), pp.~5875--5884.

\bibitem[JAF16]{johnsonvgg16}
\textsc{Johnson J., Alahi A., Fei{-}Fei L.}:
\newblock Perceptual losses for real-time style transfer and super-resolution.
\newblock \emph{CoRR abs/1603.08155} (2016).

\bibitem[KB04]{kobbelt2004survey}
\textsc{Kobbelt L., Botsch M.}:
\newblock A survey of point-based techniques in computer graphics.
\newblock \emph{Computers \& Graphics 28}, 6 (2004), 801--814.

\bibitem[KD23]{kopanas2023improving}
\textsc{Kopanas G., Drettakis G.}:
\newblock {Improving NeRF Quality by Progressive Camera Placement for
  Free-Viewpoint Navigation}.
\newblock In \emph{Vision, Modeling, and Visualization} (2023), Guthe M.,
  Grosch T., (Eds.), The Eurographics Association.

\bibitem[KKLD23]{kerbl20233d}
\textsc{Kerbl B., Kopanas G., Leimk{\"u}hler T., Drettakis G.}:
\newblock 3d gaussian splatting for real-time radiance field rendering.
\newblock \emph{ACM Transactions on Graphics 42}, 4 (2023).

\bibitem[KLL{\etalchar{*}}13]{keller2013realtime}
\textsc{Keller M., Lefloch D., Lambers M., Izadi S., Weyrich T., Kolb A.}:
\newblock Real-time {3D} reconstruction in dynamic scenes using point-based
  fusion.
\newblock In \emph{Proc. of Joint 3DIM/3DPVT Conference (3DV)} (June 2013),
  pp.~1--8.
\newblock Selected for oral presentation.

\bibitem[KLR{\etalchar{*}}22]{kopanas2022neural}
\textsc{Kopanas G., Leimk{\"u}hler T., Rainer G., Jambon C., Drettakis G.}:
\newblock Neural point catacaustics for novel-view synthesis of reflections.
\newblock \emph{ACM Transactions on Graphics (TOG) 41}, 6 (2022), 1--15.

\bibitem[KPLD21]{kopanas2021point}
\textsc{Kopanas G., Philip J., Leimk{\"u}hler T., Drettakis G.}:
\newblock Point-based neural rendering with per-view optimization.
\newblock In \emph{Computer Graphics Forum} (2021), vol.~40, Wiley Online
  Library, pp.~29--43.

\bibitem[KPZK17]{Knapitsch2017}
\textsc{Knapitsch A., Park J., Zhou Q.-Y., Koltun V.}:
\newblock Tanks and temples: Benchmarking large-scale scene reconstruction.
\newblock \emph{ACM Transactions on Graphics 36}, 4 (2017).

\bibitem[LXG22]{liao2022kitti}
\textsc{Liao Y., Xie J., Geiger A.}:
\newblock Kitti-360: A novel dataset and benchmarks for urban scene
  understanding in 2d and 3d.
\newblock \emph{IEEE Transactions on Pattern Analysis and Machine Intelligence
  45}, 3 (2022), 3292--3310.

\bibitem[LZ21]{Lassner_2021_CVPR}
\textsc{Lassner C., Zollh{\"o}fer M.}:
\newblock Pulsar: Efficient sphere-based neural rendering.
\newblock In \emph{Proceedings of the IEEE/CVF Conference on Computer Vision
  and Pattern Recognition (CVPR)} (June 2021), pp.~1440--1449.

\bibitem[MBRS{\etalchar{*}}21]{martin2021nerf}
\textsc{Martin-Brualla R., Radwan N., Sajjadi M.~S., Barron J.~T., Dosovitskiy
  A., Duckworth D.}:
\newblock Nerf in the wild: Neural radiance fields for unconstrained photo
  collections.
\newblock In \emph{Proceedings of the IEEE/CVF Conference on Computer Vision
  and Pattern Recognition} (2021), pp.~7210--7219.

\bibitem[MESK22]{muller2022instant}
\textsc{M{\"u}ller T., Evans A., Schied C., Keller A.}:
\newblock Instant neural graphics primitives with a multiresolution hash
  encoding.
\newblock \emph{arXiv preprint arXiv:2201.05989} (2022).

\bibitem[MGK{\etalchar{*}}19]{meshry2019neural}
\textsc{Meshry M., Goldman D.~B., Khamis S., Hoppe H., Pandey R., Snavely N.,
  Martin-Brualla R.}:
\newblock Neural rerendering in the wild.
\newblock In \emph{Proceedings of the IEEE/CVF Conference on Computer Vision
  and Pattern Recognition} (2019), pp.~6878--6887.

\bibitem[MKC07]{marroquim2007efficient}
\textsc{Marroquim R., Kraus M., Cavalcanti P.~R.}:
\newblock Efficient point-based rendering using image reconstruction.
\newblock In \emph{PBG@ Eurographics} (2007), pp.~101--108.

\bibitem[MSOC{\etalchar{*}}19]{mildenhall2019local}
\textsc{Mildenhall B., Srinivasan P.~P., Ortiz-Cayon R., Kalantari N.~K.,
  Ramamoorthi R., Ng R., Kar A.}:
\newblock Local light field fusion: Practical view synthesis with prescriptive
  sampling guidelines.
\newblock \emph{ACM Transactions on Graphics (TOG) 38}, 4 (2019), 1--14.

\bibitem[MST{\etalchar{*}}21]{mildenhall2021nerf}
\textsc{Mildenhall B., Srinivasan P.~P., Tancik M., Barron J.~T., Ramamoorthi
  R., Ng R.}:
\newblock Nerf: Representing scenes as neural radiance fields for view
  synthesis.
\newblock \emph{Communications of the ACM 65}, 1 (2021), 99--106.

\bibitem[M{\"u}l21]{tiny-cuda-nn}
\textsc{M{\"u}ller T.}:
\newblock {tiny-cuda-nn}, 4 2021.
\newblock URL: \url{https://github.com/NVlabs/tiny-cuda-nn}.

\bibitem[NSP{\etalchar{*}}21]{neff2021donerf}
\textsc{Neff T., Stadlbauer P., Parger M., Kurz A., Mueller J.~H., Chaitanya C.
  R.~A., Kaplanyan A., Steinberger M.}:
\newblock Donerf: Towards real-time rendering of compact neural radiance fields
  using depth oracle networks.
\newblock In \emph{Computer Graphics Forum} (2021), vol.~40, Wiley Online
  Library, pp.~45--59.

\bibitem[OLN{\etalchar{*}}22]{ost2022neural}
\textsc{Ost J., Laradji I., Newell A., Bahat Y., Heide F.}:
\newblock Neural point light fields.
\newblock In \emph{Proceedings of the IEEE/CVF Conference on Computer Vision
  and Pattern Recognition} (2022), pp.~18419--18429.

\bibitem[PGA11]{pintus2011real}
\textsc{Pintus R., Gobbetti E., Agus M.}:
\newblock Real-time rendering of massive unstructured raw point clouds using
  screen-space operators.
\newblock In \emph{Proceedings of the 12th International conference on Virtual
  Reality, Archaeology and Cultural Heritage} (2011), pp.~105--112.

\bibitem[PZ17]{penner2017soft}
\textsc{Penner E., Zhang L.}:
\newblock Soft 3d reconstruction for view synthesis.
\newblock \emph{ACM Transactions on Graphics (TOG) 36}, 6 (2017), 1--11.

\bibitem[PZVBG00]{pfister2000surfels}
\textsc{Pfister H., Zwicker M., Van~Baar J., Gross M.}:
\newblock Surfels: Surface elements as rendering primitives.
\newblock In \emph{Proceedings of the 27th annual conference on Computer
  graphics and interactive techniques} (2000), pp.~335--342.

\bibitem[RALB22]{rakhimov2022npbg++}
\textsc{Rakhimov R., Ardelean A.-T., Lempitsky V., Burnaev E.}:
\newblock {NPBG++:} accelerating neural point-based graphics.
\newblock In \emph{Proceedings of the IEEE/CVF Conference on Computer Vision
  and Pattern Recognition} (June 2022), pp.~15969--15979.

\bibitem[RFS22]{ruckert2022adop}
\textsc{R{\"u}ckert D., Franke L., Stamminger M.}:
\newblock Adop: Approximate differentiable one-pixel point rendering.
\newblock \emph{ACM Transactions on Graphics (TOG) 41}, 4 (2022), 1--14.

\bibitem[RK20]{riegler2020free}
\textsc{Riegler G., Koltun V.}:
\newblock Free view synthesis.
\newblock In \emph{European Conference on Computer Vision} (2020), Springer,
  pp.~623--640.

\bibitem[RK21]{riegler2021stable}
\textsc{Riegler G., Koltun V.}:
\newblock Stable view synthesis.
\newblock In \emph{Proceedings of the IEEE/CVF Conference on Computer Vision
  and Pattern Recognition} (2021), pp.~12216--12225.

\bibitem[RWL{\etalchar{*}}22]{ruckert2022neat}
\textsc{R\"{u}ckert D., Wang Y., Li R., Idoughi R., Heidrich W.}:
\newblock Neat: Neural adaptive tomography.
\newblock \emph{ACM Trans. Graph. 41}, 4 (2022).

\bibitem[SCCL20]{song2020deep}
\textsc{Song Z., Chen W., Campbell D., Li H.}:
\newblock Deep novel view synthesis from colored 3d point clouds.
\newblock In \emph{European Conference on Computer Vision} (2020), Springer,
  pp.~1--17.

\bibitem[SF16]{schonberger2016structure}
\textsc{Schonberger J.~L., Frahm J.-M.}:
\newblock Structure-from-motion revisited.
\newblock In \emph{Proceedings of the IEEE conference on computer vision and
  pattern recognition} (2016), pp.~4104--4113.

\bibitem[SK00]{shum2000review}
\textsc{Shum H., Kang S.~B.}:
\newblock Review of image-based rendering techniques.
\newblock In \emph{Visual Communications and Image Processing 2000} (2000),
  vol.~4067, SPIE, pp.~2--13.

\bibitem[SKW19]{schutz2019real}
\textsc{Sch{\"u}tz M., Kr{\"o}sl K., Wimmer M.}:
\newblock Real-time continuous level of detail rendering of point clouds.
\newblock In \emph{2019 IEEE Conference on Virtual Reality and 3D User
  Interfaces (VR)} (2019), IEEE, pp.~103--110.

\bibitem[SKW21]{schutz2021rendering}
\textsc{Sch{\"u}tz M., Kerbl B., Wimmer M.}:
\newblock Rendering point clouds with compute shaders and vertex order
  optimization.
\newblock In \emph{Computer Graphics Forum} (2021), vol.~40, Wiley Online
  Library, pp.~115--126.

\bibitem[SKW22]{schutz2022software}
\textsc{Sch{\"u}tz M., Kerbl B., Wimmer M.}:
\newblock Software rasterization of 2 billion points in real time.
\newblock \emph{arXiv preprint arXiv:2204.01287} (2022).

\bibitem[SMB{\etalchar{*}}20]{sitzmann2020implicit}
\textsc{Sitzmann V., Martel J., Bergman A., Lindell D., Wetzstein G.}:
\newblock Implicit neural representations with periodic activation functions.
\newblock \emph{Advances in Neural Information Processing Systems 33} (2020).

\bibitem[SSS06]{snavely2006photo}
\textsc{Snavely N., Seitz S.~M., Szeliski R.}:
\newblock Photo tourism: exploring photo collections in 3d.
\newblock In \emph{ACM siggraph 2006 papers}. 2006, pp.~835--846.

\bibitem[STB{\etalchar{*}}19]{srinivasan2019pushing}
\textsc{Srinivasan P.~P., Tucker R., Barron J.~T., Ramamoorthi R., Ng R.,
  Snavely N.}:
\newblock Pushing the boundaries of view extrapolation with multiplane images.
\newblock In \emph{Proceedings of the IEEE/CVF Conference on Computer Vision
  and Pattern Recognition} (2019), pp.~175--184.

\bibitem[STH{\etalchar{*}}19]{sitzmann2019deepvoxels}
\textsc{Sitzmann V., Thies J., Heide F., Nie{\ss}ner M., Wetzstein G.,
  Zollhofer M.}:
\newblock Deepvoxels: Learning persistent 3d feature embeddings.
\newblock In \emph{Proceedings of the IEEE/CVF Conference on Computer Vision
  and Pattern Recognition} (2019), pp.~2437--2446.

\bibitem[SZPF16]{schoenberger2016mvs}
\textsc{Sch\"{o}nberger J.~L., Zheng E., Pollefeys M., Frahm J.-M.}:
\newblock Pixelwise view selection for unstructured multi-view stereo.
\newblock In \emph{European Conference on Computer Vision (ECCV)} (2016).

\bibitem[TMW{\etalchar{*}}21]{tancik2021learned}
\textsc{Tancik M., Mildenhall B., Wang T., Schmidt D., Srinivasan P.~P., Barron
  J.~T., Ng R.}:
\newblock Learned initializations for optimizing coordinate-based neural
  representations.
\newblock In \emph{Proceedings of the IEEE/CVF Conference on Computer Vision
  and Pattern Recognition} (2021), pp.~2846--2855.

\bibitem[TRS22]{turki2022mega}
\textsc{Turki H., Ramanan D., Satyanarayanan M.}:
\newblock Mega-nerf: Scalable construction of large-scale nerfs for virtual
  fly-throughs.
\newblock In \emph{Proceedings of the IEEE/CVF Conference on Computer Vision
  and Pattern Recognition} (2022), pp.~12922--12931.

\bibitem[TS20]{tucker2020single}
\textsc{Tucker R., Snavely N.}:
\newblock Single-view view synthesis with multiplane images.
\newblock In \emph{Proceedings of the IEEE/CVF Conference on Computer Vision
  and Pattern Recognition} (2020), pp.~551--560.

\bibitem[TTM{\etalchar{*}}22]{tewari2022advances}
\textsc{Tewari A., Thies J., Mildenhall B., Srinivasan P., Tretschk E., Yifan
  W., Lassner C., Sitzmann V., Martin-Brualla R., Lombardi S., et~al.}:
\newblock Advances in neural rendering.
\newblock In \emph{Computer Graphics Forum} (2022), vol.~41, Wiley Online
  Library, pp.~703--735.

\bibitem[TZN19]{thies2019deferred}
\textsc{Thies J., Zollh{\"o}fer M., Nie{\ss}ner M.}:
\newblock Deferred neural rendering: Image synthesis using neural textures.
\newblock \emph{ACM Transactions on Graphics (TOG) 38}, 4 (2019), 1--12.

\bibitem[VVP20]{vasilakis2020survey}
\textsc{Vasilakis A.-A., Vardis K., Papaioannou G.}:
\newblock A survey of multifragment rendering.
\newblock In \emph{Computer Graphics Forum} (2020), vol.~39, Wiley Online
  Library, pp.~623--642.

\bibitem[WGSJ20]{wiles2020synsin}
\textsc{Wiles O., Gkioxari G., Szeliski R., Johnson J.}:
\newblock Synsin: End-to-end view synthesis from a single image.
\newblock In \emph{Proceedings of the IEEE/CVF Conference on Computer Vision
  and Pattern Recognition} (2020), pp.~7467--7477.

\bibitem[WSMG{\etalchar{*}}16]{whelan2016elasticfusion}
\textsc{Whelan T., Salas-Moreno R.~F., Glocker B., Davison A.~J., Leutenegger
  S.}:
\newblock Elasticfusion: Real-time dense slam and light source estimation.
\newblock \emph{The International Journal of Robotics Research 35}, 14 (2016),
  1697--1716.

\bibitem[XXP{\etalchar{*}}22]{xu2022point}
\textsc{Xu Q., Xu Z., Philip J., Bi S., Shu Z., Sunkavalli K., Neumann U.}:
\newblock Point-nerf: Point-based neural radiance fields.
\newblock In \emph{Proceedings of the IEEE/CVF Conference on Computer Vision
  and Pattern Recognition} (2022), pp.~5438--5448.

\bibitem[YCA{\etalchar{*}}20]{Yang_2020_CVPR}
\textsc{Yang Z., Chai Y., Anguelov D., Zhou Y., Sun P., Erhan D., Rafferty S.,
  Kretzschmar H.}:
\newblock Surfelgan: Synthesizing realistic sensor data for autonomous driving.
\newblock In \emph{Proceedings of the IEEE/CVF Conference on Computer Vision
  and Pattern Recognition (CVPR)} (June 2020).

\bibitem[YLT{\etalchar{*}}21]{yu2021plenoctrees}
\textsc{Yu A., Li R., Tancik M., Li H., Ng R., Kanazawa A.}:
\newblock Plenoctrees for real-time rendering of neural radiance fields.
\newblock In \emph{Proceedings of the IEEE/CVF International Conference on
  Computer Vision} (2021), pp.~5752--5761.

\bibitem[YLY{\etalchar{*}}19]{yu2019free}
\textsc{Yu J., Lin Z., Yang J., Shen X., Lu X., Huang T.~S.}:
\newblock Free-form image inpainting with gated convolution.
\newblock In \emph{Proceedings of the IEEE/CVF international conference on
  computer vision} (2019), pp.~4471--4480.

\bibitem[YSW{\etalchar{*}}19]{yifan2019differentiable}
\textsc{Yifan W., Serena F., Wu S., {\"O}ztireli C., Sorkine-Hornung O.}:
\newblock Differentiable surface splatting for point-based geometry processing.
\newblock \emph{ACM Transactions on Graphics (TOG) 38}, 6 (2019), 1--14.

\bibitem[YYTK21]{yu2021pixelnerf}
\textsc{Yu A., Ye V., Tancik M., Kanazawa A.}:
\newblock pixelnerf: Neural radiance fields from one or few images.
\newblock In \emph{Proceedings of the IEEE/CVF Conference on Computer Vision
  and Pattern Recognition} (2021), pp.~4578--4587.

\bibitem[ZBRH22]{zhang2022differentiable}
\textsc{Zhang Q., Baek S.-H., Rusinkiewicz S., Heide F.}:
\newblock Differentiable point-based radiance fields for efficient view
  synthesis.
\newblock \emph{arXiv preprint arXiv:2205.14330} (2022).

\bibitem[ZIE{\etalchar{*}}18]{zhang2018perceptual}
\textsc{Zhang R., Isola P., Efros A.~A., Shechtman E., Wang O.}:
\newblock The unreasonable effectiveness of deep features as a perceptual
  metric.
\newblock In \emph{CVPR} (2018).

\bibitem[ZPVBG01]{zwicker2001surface}
\textsc{Zwicker M., Pfister H., Van~Baar J., Gross M.}:
\newblock Surface splatting.
\newblock In \emph{Proceedings of the 28th annual conference on Computer
  graphics and interactive techniques} (2001), pp.~371--378.

\bibitem[ZTF{\etalchar{*}}18]{zhou2018stereo}
\textsc{Zhou T., Tucker R., Flynn J., Fyffe G., Snavely N.}:
\newblock Stereo magnification: Learning view synthesis using multiplane
  images.
\newblock \emph{ACM Transactions on Graphics (TOG) 37}, 4 (2018), 1--12.

\bibitem[ZTS{\etalchar{*}}16]{zhou2016view}
\textsc{Zhou T., Tulsiani S., Sun W., Malik J., Efros A.~A.}:
\newblock View synthesis by appearance flow.
\newblock In \emph{European conference on computer vision} (2016), Springer,
  pp.~286--301.

\end{thebibliography}

\newpage
\setcounter{section}{0}
\def\thesection{\Alph{section}}
\section{Individual Tabs: MipNeRF-360 (MipNeRF-360 resolutions)}
	\begin{table}[h]
	\caption{LPIPS\textsubscript{VGG} scores for Mip-NeRF360 scenes.  $\dagger$ copied from original paper~\cite{barron2022mip}. $ \ddagger $ copied from 3D GS~\cite{kerbl20233d}. Image resolutions as in MipNerf-360: half resolution for indoor, quarter resolution for outdoor. Average ours: 0.176}
	\scalebox{0.6}{
		\centering
    \begin{tabular}{l|ccccc|cccc}
	~ & bicycle & flowers & garden & stump & treehill  & room & counter & kitchen & bonsai \\ \hline
	InstantNGP$ \ddagger $ & 0.446 & 0.441 & 0.257 & 0.421 & 0.450  & 0.261 & 0.306 & 0.195 & 0.205 \\ 
	Mip-NeRF 360$^\dagger$ & 0.301 & 0.344 & 0.170 & 0.261 &  0.339 & {0.211} & {0.204} & {0.127} & {0.176} \\ 
 	Mip-NeRF 360$ \ddagger $ & 0.305 & 0.346 & 0.171 & 0.265 & 0.347 & 0.213 & 0.207 & 0.128 & 0.179\\
	Gaussian Spl.$ \ddagger $ & {0.205} & {0.336} & {0.103} & {0.210} & {0.317} & 0.220 & {0.204} & 0.129 & 0.205 \\ 
    TRIPS(ours) & 0.194 & 0.297 & 0.159 & 0.268 & 0.266 & 0.147 & 0.158 & 0.127 & 0.111   \\
    \end{tabular}
	}	\label{tab:360_scene_lpips}
	\end{table}
\begin{table}[h]
	\caption{Normalized LPIPS\textsubscript{VGG} scores: based on the original paper~\cite{zhang2018perceptual}, images should be normalized between -1 and 1 (as is in every table except Appendix Tab.~\ref{tab:360_scene_lpips}). Scored of ours with this normalization. Average ours: 0.213}
	\scalebox{0.6}{
		\centering
    \begin{tabular}{l|ccccc|cccc}
	~ & bicycle & flowers & garden & stump & treehill  & room & counter & kitchen & bonsai \\ \hline
    TRIPS(ours) & 0.223 & 0.318 & 0.183 & 0.309 & 0.308 & 0.197 & 0.206 & 0.154 & 0.153  \\
    \end{tabular}
    }	\label{tab:360_scene_lpips_fix}

\end{table}
 \begin{table}[h]
	\caption{PSNR scores for Mip-NeRF360 scenes. $\dagger$ copied from original paper~\cite{barron2022mip}. $ \ddagger $ copied from Kerbl and Kopanas et al.~\cite{kerbl20233d}. Image resolutions as in MipNerf-360: half resolution for indoor, quarter resolution for outdoor. Average ours: 25.94}
	\scalebox{0.6}{
		\centering
		\centering
		\begin{tabular}{l|ccccc|cccc}
			~ & bicycle & flowers & garden & stump & treehill  & room & counter & kitchen & bonsai \\ \hline
			InstantNGP$ \ddagger $ & 22.171 & 20.652 & 25.069 & 23.466 & 22.373  & 29.690 & 26.691 & 29.479 & 30.685 \\ 
			Mip-NeRF 360$^\dagger$ & 24.37 & {21.73} & 26.98 & 26.40 & 22.87 & {31.63} & {29.55} & {32.23} & {33.46} \\
   			Mip-NeRF 360$ \ddagger $ & 24.305 & 21.649 & 26.875 & 26.175 & {22.929} & 31.467 & 29.447 & 31.989 & 33.397 \\
			Gaussian Spl.$ \ddagger $  & {25.246} & 21.520 & {27.410} & {26.550} & 22.490 & 30.632 & 28.700 & 30.317 & 31.980 \\ 
            TRIPS(ours) & 23.466 & 19.439 & 25.384 & 24.174 & 22.044 & 29.066 & 27.002 & 27.662 & 28.710 \\

		\end{tabular}
	}
	\label{tab:360_scene_psnr}
\end{table}
\begin{table}[h!]

	\caption{SSIM scores for Mip-NeRF360 scenes. $\dagger$ copied from original paper~\cite{barron2022mip}. $ \ddagger $ copied from Kerbl and Kopanas et al.~\cite{kerbl20233d}. Image resolutions as in MipNerf-360: half resolution for indoor, quarter resolution for outdoor. Average ours: 0.778}
	\scalebox{0.6}{
		\centering
	    \begin{tabular}{l|ccccc|cccc}
		~ & bicycle & flowers & garden & stump & treehill  & room & counter & kitchen & bonsai \\ \hline
		InstantNGP$ \ddagger $ & 0.512 & 0.486 & 0.701 & 0.594 & 0.542  & 0.871 & 0.817 & 0.858 & 0.906 \\ 
		Mip-NeRF 360$^\dagger$ & 0.685 & 0.583 & 0.813 & 0.744 & 0.632 & 0.913 & 0.894 & 0.920 & {0.941} \\ 
		Mip-NeRF 360$ \ddagger $ & 0.685 & 0.584 & 0.809 & 0.745 & 0.631 & 0.910 & 0.892 & 0.917 & 0.938\\
		Gaussian Spl.$ \ddagger $ & {0.771} & {0.605} & {0.868} & {0.775} & {0.638} & {0.914} & {0.905} & {0.922} & 0.938 \\ 
        TRIPS(ours) & 0.704 & 0.502 & 0.773 & 0.681 & 0.591 & 0.883 & 0.845 & 0.850 & 0.899 \\
	\end{tabular}
	}
	\label{tab:360_scene_ssim}
\end{table}
\newpage

\section{Individual Tabs: Tanks and Temples}

\begin{table}[h]
	\caption{LPIPS\textsubscript{VGG} scores for Tanks and Temples scenes (intermediate set).}
	\scalebox{0.6}{
    \centering
    \begin{tabular}{l|cccccccc}
    ~ & playground & lighthouse & francis & m60 & train & panther & family & horse \\ \hline
    InstantNGP       & 0.581 & 0.477 & 0.472 & 0.414 & 0.527 & 0.410 & 0.456 & 0.437 \\
    Mip-NeRF 360    & 0.350 & 0.346 & 0.343 & 0.313 & 0.486 & 0.285 & 0.277 & 0.244 \\
    Gaussian Spl.   & 0.322 & 0.296 & 0.345 & 0.273 & 0.344 & 0.267 & 0.262 & 0.244 \\
    ADOP            & 0.233 & 0.210 & 0.241 & 0.226 & 0.239 & 0.232 & 0.225 & 0.216 \\
    TRIPS(ours)     & 0.229 & 0.208 & 0.221 & 0.208 & 0.223 & 0.207 & 0.202 & 0.194 \\
    \end{tabular}
	}	
    \label{tab:tt_scene_lpips} 
\end{table}

\begin{table}[h]
	\caption{PSNR scores for Tanks and Temples scenes (intermediate set).}
	\scalebox{0.6}{
    \centering
    \begin{tabular}{l|cccccccc}
    ~ & playground & lighthouse & francis & m60 & train & panther & family & horse \\ \hline
InstantNGP        & 18.224 & 20.783 & 23.148 & 24.115 & 18.753 & 26.312 & 21.453 & 19.719 \\
Mip-NeRF 360      & 25.200 & 22.379 & 28.266 & 24.743 & 18.674 & 27.428 & 25.326 & 25.659 \\
Gaussian Spl.    & 24.611 & 21.592 & 25.993 & 26.972 & 20.990 & 27.823 & 24.491 & 23.880 \\
ADOP             & 24.856 & 23.057 & 22.036 & 24.707 & 22.335 & 25.666 & 24.013 & 23.261 \\
TRIPS(ours)      & 25.116 & 23.382 & 24.818 & 25.832 & 22.974 & 26.841 & 23.532 & 23.174 \\
    \end{tabular}
	}	
    \label{tab:tt_scene_psnr} 
\end{table}

\begin{table}[h]
	\caption{SSIM scores for Tanks and Temples scenes (intermediate set).}
	\scalebox{0.6}{
    \centering
    \begin{tabular}{l|cccccccc}
    ~ & playground & lighthouse & francis & m60 & train & panther & family & horse \\ \hline
    InstantNGP & 0.493 & 0.713 & 0.764 & 0.766 & 0.596 & 0.808 & 0.681 & 0.721 \\
    Mip-NeRF 360  & 0.741 & 0.771 & 0.847 & 0.837 & 0.619 & 0.863 & 0.815 & 0.858 \\
    Gaussian Spl. & 0.766 & 0.790 & 0.847 & 0.868 & 0.734 & 0.880 & 0.820 & 0.853 \\
    ADOP   & 0.753 & 0.796 & 0.827 & 0.843 & 0.755 & 0.844 & 0.775 & 0.817 \\
    TRIPS(ours)       & 0.746 & 0.787 & 0.848 & 0.849 & 0.764 & 0.851 & 0.791 & 0.825 \\

    \end{tabular}
	}	
    \label{tab:tt_scene_ssim} 
\end{table}

\newpage

\section{Individual Tabs: MipNeRF-360 (our resolutions)}

\begin{table}[h]
	\caption{LPIPS\textsubscript{VGG} scores for MipNeRF-360 scenes with our resolutions (half indoor and outdoor).}
	\scalebox{0.6}{
    \centering
	    \begin{tabular}{l|ccccc|cccc}
		~ & bicycle & flowers & garden & stump & treehill  & room & counter & kitchen & bonsai \\ \hline
InstantNGP      & 0.600 & 0.587 & 0.516 & 0.591 & 0.606 & 0.354 & 0.393 & 0.286 & 0.266 \\
Mip-NeRF 360    & 0.423 & 0.458 & 0.310 & 0.385 & 0.460 & 0.223 & 0.238 & 0.162 & 0.169 \\
Gaussian Spl    & 0.363 & 0.448 & 0.245 & 0.359 & 0.460 & 0.234 & 0.231 & 0.158 & 0.215 \\
ADOP            & 0.319 & 0.409 & 0.259 & 0.376 & 0.422 & 0.241 & 0.264 & 0.221 & 0.223 \\
TRIPS(ours)     & 0.284 & 0.383 & 0.219 & 0.327 & 0.358 & 0.197 & 0.206 & 0.154 & 0.153 \\
    \end{tabular}
	}	
    \label{tab:own369_scene_lpips} 
\end{table}

\begin{table}[h]
	\caption{PSNR scores for MipNeRF-360 scenes with our resolutions (half indoor and outdoor).}
	\scalebox{0.6}{
    \centering
	    \begin{tabular}{l|ccccc|cccc}
		~ & bicycle & flowers & garden & stump & treehill  & room & counter & kitchen & bonsai \\ \hline
InstantNGP & 21.479 & 19.880 & 23.556 & 22.791 & 21.828 & 29.347 & 26.618 & 28.528 & 30.904 \\
Mip-NeRF 360 & 23.541 & 21.082 & 25.887 & 26.219 & 22.525 & 31.711 & 29.425 & 31.351 & 33.222 \\
Gaussian Spl. & 24.286 & 20.732 & 25.690 & 26.123 & 22.274 & 30.423 & 28.987 & 30.446 & 27.225\\ 
ADOP & 21.910 & 19.432 & 23.711 & 23.700 & 20.312 & 25.975 & 23.088 & 23.614 & 24.330 \\
TRIPS(ours)  & 22.961 & 19.668 & 25.385 & 24.964 & 21.725 & 29.066 & 27.002 & 27.662 & 28.710 \\
    \end{tabular}
	}	
    \label{tab:own369_scene_psnr} 
\end{table}

\begin{table}[h]
	\caption{SSIM scores for MipNeRF-360 scenes with our resolutions (half indoor and outdoor).}
	\scalebox{0.6}{
    \centering
	    \begin{tabular}{l|ccccc|cccc}
		~ & bicycle & flowers & garden & stump & treehill  & room & counter & kitchen & bonsai \\ \hline
InstantNGP & 0.486 & 0.422 & 0.545 & 0.568 & 0.524 & 0.853 & 0.789 & 0.811 & 0.895 \\
Mip-NeRF 360 & 0.635 & 0.522 & 0.730 & 0.727 & 0.611 & 0.909 & 0.882 & 0.901 & 0.940 \\
Gaussian Spl. & 0.693 & 0.530 & 0.764 & 0.748 & 0.600 & 0.896 & 0.892 & 0.899 & 0.853\\ 
ADOP & 0.610 & 0.475 & 0.674 & 0.652 & 0.546 & 0.839 & 0.769 & 0.737 & 0.818 \\
TRIPS(ours)  & 0.668 & 0.482 & 0.751 & 0.707 & 0.587 & 0.883 & 0.845 & 0.850 & 0.899 \\
    \end{tabular}
	}	
    \label{tab:own369_scene_ssim} 
\end{table}

\newpage

\end{document}